\documentclass[10pt]{article}
\usepackage[legalpaper, margin=1in]{geometry}

\usepackage{graphicx}

\usepackage{tikz}
\usepackage{comment}
\usepackage{amsmath,amssymb} 
\usepackage{wrapfig}
\usepackage{hyperref}
\usepackage{amsthm}

\usepackage{booktabs}
\usepackage{enumitem}
\usepackage{subcaption}
\usepackage{bbm}
\usepackage{multirow}

\newtheorem{definition}{Definition}

\newtheorem{theorem}{Theorem}

\def\BX{\mathbf{X}}
\def\CX{\mathcal{X}}
\def\bx{\mathbf{x}}
\def\CT{\mathcal{T}}

\begin{document}

\title{Theoretical Understanding of the  Information Flow on Continual Learning Performance} 
\author{Josh Andle,  Salimeh Yasaei Sekeh}%

\date{}
\maketitle

\begin{abstract}
Continual learning (CL) is a setting in which an agent has to learn from an incoming stream of data sequentially. CL performance evaluates the model's ability to continually learn and solve new problems with incremental available information over time while retaining previous knowledge. Despite the numerous previous solutions to bypass the catastrophic forgetting (CF) of previously seen tasks during the learning process, most of them still suffer significant forgetting, expensive memory cost, or lack of theoretical understanding of neural networks' conduct while learning new tasks. While the issue that CL performance degrades under different training regimes has been extensively studied empirically, insufficient attention has been paid from a theoretical angle.  In this paper, we establish a probabilistic framework to analyze information flow through layers in networks for task sequences and its impact on learning performance. Our objective is to optimize the {\it information preservation between layers} while learning new tasks to manage task-specific knowledge passing throughout the layers while maintaining model performance on previous tasks. In particular, we study CL performance's relationship with information flow in the network to answer the question {\it How can knowledge of information flow between layers be used to alleviate CF?} Our analysis provides novel insights of information adaptation within the layers during the incremental task learning process.
Through our experiments, we provide empirical evidence and practically highlight the performance improvement across multiple tasks.  

\end{abstract}

\section{Introduction}
Humans are continual learning systems that have been very successful at adapting to new situations while not forgetting about their past experiences. Similar to the human brain, continual learning (CL) tackles the setting of learning new tasks sequentially without dismissing information learned from the previous tasks and no earlier learned concepts are forgotten~\cite{chen2018lifelong,li2017learning,parisi2019continual}. 
A wide variety of CL methods mainly either minimize a loss function which is a combination of forgetting and generalization loss to overcome catastrophic forgetting~\cite{ke2020continual,kirkpatrick2017overcoming,mirzadeh2020understanding,yin2020optimization,yoon2017lifelong} or improve quick generalization~\cite{finn2017model,vinyals2016matching}. While these approaches have demonstrated state-of-the-art performance and achieve some degree of continual learning in deep neural networks, there has been limited prior work extensively and analytically investigating the impact that different training regimes can have on learning a sequence of tasks. Although major advances have been made in the field, one recurring problem that still remains not completely solved is that of catastrophic forgetting (CF). An approach to address this goal is to gradually extend acquired knowledge learned within layers in the network and use it for future learning. While the CF issue has been extensively studied empirically, little attention has been paid from a theoretical angle~\cite{jung2020continual,mirzadeh2020understanding,raghavan2021formalizing}. To the best of our knowledge, there are no works which explain what occurs when certain portions of a network are more important than others for passing information of a given task downstream to the end of the network.
In this paper, we explore the CL performance and CF problem from a probabilistic perspective. We seek to understand the connection of the passing of information downstream through layers in the network and learning performance at a more in-depth and fundamental theoretical level.  We integrate these studies into two central questions: 
\begin{itemize}
    \item[(1)] {\it Given a sequence of joint random variables and tasks, how much does information flow between layers influence the learning performance and alleviate CF?}
    \item[(2)] {\it Given a sequence of tasks, how much does the sparsity level of layers on task-specific training influence the forgetting? }
\end{itemize}
The answers to these questions are theoretically and practically important for continual learning research because: (1) despite the tangible improvements in task learning, the core problem of deep network efficiency on performance assists selective knowledge sharing through downstream information within layers; (2) a systematic understanding of learning tasks provides schemes to accommodate more tasks to learn; and (3) monitoring information flow in the network for each task alleviates forgetting.\\
Toward our analysis, we measure the information flow between layers given a task by using dependency measures between filters in consecutive layers conditioned on tasks. Given a sequence of joint random variables and tasks, we compute the forgetting by the correlation between task and trained model's loss on the tasks in the sequence.
\noindent To summarize, our contributions in this paper are,
\begin{itemize}
    \item Introducing the new concept of {\it task-sensitivity}, which targets task-specific knowledge passing through layers in the network.
    \item Providing a theoretical and in-depth analysis of information flow through layers in networks for task sequences and its impact on learning performance.
    \item Optimizing the information preservation between layers while learning new tasks by freezing task-specific important filters.
    \item Developing a new bound on expected forgetting using optimal freezing mask.
    \item Providing experimental evidence and practical observations of layer connectivities in the network and their impact on accuracy. 
\end{itemize}

\paragraph{Organization:}
The paper is organized as follows. In Section~\ref{probelm.formulation} we briefly review the continual learning problem formulation and fundamental definitions of the performance. In addition, a set of new concepts including task sensitivity and task usefulness of layers in the network is introduced. In Section~\ref{sec.CL.Performance} we establish a series of foundational and theoretical findings that focus on performance and forgetting analysis in terms of the sensitivity property of layers.  A new bound on expected forgetting based on the optimal freezing mask is given in this section. Finally, in Section~\ref{section.experiment} we provide experimental evidence of our analysis using the CIFAR-10/100 and Permuted MNIST datasets. 
The main proofs of the theorems in the paper are given in the supplementary materials, although in Section~\ref{Sec.key.com} we provide the key components and techniques that we use for the proofs. 
\section{Problem Formulation}\label{probelm.formulation}
In supervised continual learning, we are given a sequence of joint random variables $(\BX_t, T_t)$, with realization space $\CX_t\times\CT_t$ where $(\bx_t,y_t)$ is an instance of the $\CX_t\times\CT_t$ space. We use $\|.\|$ to denote the Euclidean norm for vectors and $\|.\|_F$ to denote the Frobenius norm for matrices. In this section, we begin by presenting a brief list of notations and then provide the key definitions.\\ 
\noindent{\it Notations:}
We assume that a given DNN has a total of $L$ layers where,\vspace{-0.2cm}
\begin{itemize}
    \item $F^{(L)}$: A function mapping the input space $\mathcal{X}$ to a set of classes $\mathcal{T}$, i.e. $F^{(L)}: \mathcal{X}\mapsto \mathcal{T}$.
    \item $f^{(l)}$: The $l$-th layer of $F^{(L)}$ with  $M_l$ as number of filters in layer $l$.
    \item $f^{(l)}_i$: $i$-th filter in layer $l$.
    \item $F^{(i,j)}:= f^{(j)}\circ\ldots \circ f^{(i)}$: A subnetwork which is a group of consecutive layers $f^{(i)},\ldots,f^{(j)}$.
    \item  $F^{(j)}:= F^{(1,j)}=f^{(j)}\circ\ldots \circ f^{(1)}$: First part of the network up to layer $j$.
    \item  $\sigma ^{(l)}$: The activation function in layer $l$.
    \item $\widetilde{f}_{t}^{(l)}$: Sensitive layer for task $t$.
    \item $\widetilde{F}_t^{(L)}:= F^{(L)}_t/\widetilde{f}_{t}^{(l)}$: The network with $L$ layers when $l$-th sensitive layer $\widetilde{f}^{(l)}$ is frozen while training on task $t$.
    \item $\pi(T_t)$: The prior probability of class label $T_t\in \mathcal{T}_t$. 
    \item $\eta_{tl}$, $\gamma_{tl}$: Thresholds for sensitivity and usefulness of $l$-th layer $f^{(l)}$ for task $t$. 
\end{itemize}
In this section, we revisit the standard definition of training performance and forgetting and define the new concepts {\it task-sensitive layer} and {\it task-useful layer}.   
\begin{definition}\label{def.sensitivity}
{\rm (Task-Sensitive Layer)} The $l$-th layer, $f^{(l)}$, is called a {\it $t$-task-sensitive} layer if the average information flow between filters in consecutive layers $l$ and $l+1$ is high i.e.
\begin{equation}\label{Def: Delta}
  \Delta_t(f^{(l)},f^{(l+1)}):=  \frac{1}{M_{l}\; M_{l+1}}\sum\limits_{i=1}^{M_l}\sum\limits_{j=1}^{M_{l+1}} \rho\left(f^{(l)}_i,f^{(l+1)}_j|T_t\right)\geq \eta_{lt},
\end{equation}
where $\rho$ is a connectivity measure given task $T_t$ such as conditional Pearson correlation or conditional Mutual Information~\cite{TC,Csisz1967}. In this work we focus on only Pearson correlation as the connectivity measure between layers $l$ and $l+1$. 
\end{definition}
Without loss of generality, in this work we assume that filters $f_i^{(l)}$, $i=1,\ldots,M_l$, are normalized such that
$$
\mathbb{E}_{(\mathbf{X}_t,T_t)\sim D_t}\left[f_i^{(l)}(\mathbf{X}_t)|T_t\right]=0\;\;\hbox{and}\;\;\;\mathbb{V}\left[f_i^{(l)}(\mathbf{X}_t)|T_t\right]=1,\;\; \; l=1,\ldots,L,
$$
Therefore the Pearson correlation between the $i$-th filter in layer $l$ and the $j$-th filter in layer $l+1$ becomes 
\begin{equation}\label{Pearson-corr}
        \rho(f_i^{(l)}, f_j^{(l+1)}|T_t):=\mathbb{E}_{(\mathbf{X}_t,T_t)\sim D_t}\left[f_i^{(l)}(\mathbf{X}_t)f_j^{(l+1)}(\mathbf{X}_t)|T_t\right].
    \end{equation}
Note that in this paper we consider the absolute value of $\rho$ in the range $[0,1]$. 
\begin{definition} \label{def-useful}{\rm (Task-Useful Layer)} Suppose input $\mathbf{X}_t$ and task $T_t$ have joint distribution $\mathcal{D}_t$. For a given distribution $\mathcal{D}_t$, the $l$-layer $f^{(l)}$ is called {\it $t$-task-useful} if there exist two mapping functions $G_l: \mathcal{L}_l \mapsto \mathcal{T}_t$ and $K_l: \mathcal{X}_t \mapsto \mathcal{L}_l$ such that
\begin{equation}\label{semi-def:fromula}
    {\mathbb{E}}_{(\mathbf{X}_t,T_t)\sim \mathcal{D}_t}\big[T_t \cdot G_l\circ f^{(l)}(K_{l-1} \circ\mathbf{X}_t)\big]\geq \gamma_{tl}.
\end{equation}
Note that here $f^{(l)}$ is a map function $f^{(l)}: \mathcal{L}_{l-1} \mapsto \mathcal{L}_l$.
\end{definition}
Within this formulation, two parameters determine the contributions of the $l$-th layer of network $F^{(l)}$ on task $t$: $\eta_{tl}$ the contribution of passing forward the information flow to the next consecutive layer, and $\gamma_{tl}$, the contribution of the $l$-th layer in learning task $t$. 
 Training a neural network $F_t^{(L)}\in \mathcal{F}$ is performed by minimizing a loss function (empirical risk) that decreases with the correlation between the weighted combination of the networks and the label: 
\begin{equation}\label{eq-loss-func}\begin{array}{l}
\mathbb{E}_{(\mathbf{X}_t,T_t)\sim D_t}\left\{L_t(F^{(L)}_t(\BX_{t}),T_{t})\right\} \\
\qquad= - \mathbb{E}_{(\mathbf{X}_t,T_t)\sim D_t}\left\{T_t\cdot\left(b+\sum\limits_{F_t\in\mathcal{F}}w_{F_t}\cdot F_t^{(L)}(\BX_t)\right)\right\}. 
\end{array}\end{equation}
We remove offset $b$ without loss of generality. Define \begin{equation}\label{loss-func}
\ell_t(\omega):=-\sum\limits_{F_t\in\mathcal{F}}w_{F_t}\cdot F_t^{(L)}(\mathbf{X}_t),
\end{equation}
therefore the loss function in (\ref{eq-loss-func}) becomes
\begin{equation}\label{eq-loss-func2}
\mathbb{E}_{(\mathbf{X}_t,T_t)\sim D_t}\left\{L_t(F^{(L)}_t(\BX_{t}),T_{t})\right\} = \mathbb{E}_{(\mathbf{X}_t,T_t)\sim D_t}\left\{T_t\cdot \ell_t(\omega)\right\}.   
\end{equation}
Let $\omega_t^*$ be the set of parameters when the network is trained on task $T_t$ that minimizes the loss function (\ref{eq-loss-func}): 
   \begin{align}
      \omega_{t}^*:= {argmin}_{\omega_{t}} \mathbb{E}_{(\mathbf{X}_t,T_t)\sim D_t}\left\{ T_t\cdot\left(\ell_t({\omega}_{t})\right)\right\},
   \end{align}
where $\ell_t$ is defined in (\ref{loss-func}). The total risk of all seen tasks given limited or no access to data $\CX_t\times\CT_t$ from random variables $(\BX_t, T_t)$ from previous tasks $t<\tau$ i.e. $T_1,T_2,\ldots,T_\tau$ is given by 
\begin{equation}\label{total-risk}
\sum\limits_{t=1}^\tau \mathbb{E}_{(\mathbf{X}_t,T_t)\sim D_t}\left\{T_t\cdot \ell_t(\omega_\tau)\right\}.
\end{equation}
The set of parameters when the network $F^{(l)}$ is trained after seeing all tasks is the solution of minimizing the risk in (\ref{total-risk}) and is denoted by $\omega^*_\tau$. 
\begin{definition}\label{perform-diff}
{\rm (Performance Difference)} Suppose input $\mathbf{X}_t$ and task $T_t$ have joint distribution $\mathcal{D}_t$. Let $\widetilde{F}_t^{(L)}:=F^{(L)}_t/\widetilde{f}_{t}^{(l)}\in \mathcal{F}$ be the network with $L$ layers when $l$-layer $f^{(l)}$ is frozen while training on task $t$. The performance difference between training $F_t^{(l)}$ and $\widetilde{F}_t^{(L)}$ is defined as
\begin{equation}
   d(F^{(L)}_t,\widetilde{F}_t^{(L)}):=  \mathbb{E}_{(\mathbf{X}_t,T_t)\sim D_t}\left\{L_t(F^{(L)}_t(\mathbf{X}_{t}),T_{t}) - L_t(\widetilde{F}_t^{(L)}(\mathbf{X}_{t}),T_{t}) \right\}.
\end{equation}
Let ${\omega}^*_t$ and $\widetilde{\omega}^{*}_t$ be the convergent or optimum parameters after training $F^{(L)}_t$ and $\widetilde{F}_t^{(L)}$ has been finished for task $t$, respectively. The {\it training deviation} for task $t$ is defined as:
\begin{equation}\label{training-deviation}
    {\delta}_t (\omega^*_t|\widetilde{\omega}^*_t):= \ell_t(\omega^*_t)-\ell_t(\widetilde{\omega}^*_t).
\end{equation}
The optimal performance difference in Definition~\ref{perform-diff} is the average of ${\delta}_t$ in (\ref{training-deviation}): 
\begin{equation*}
d(F^{(L)}_t,\widetilde{F}_t^{(L)})=    \mathbb{E}_{(\mathbf{X}_t,T_t)\sim D_t}\left[T_t\cdot{\delta}_t (\omega^*_t|\widetilde{\omega}^*_t)\right]= \mathbb{E}_{(\mathbf{X}_t,T_t)\sim D_t}\big[T_t\cdot\big(\ell_t(\omega^*_t)-\ell_t(\widetilde{\omega}^*_t)\big)\big].
\end{equation*}
\end{definition}
\section{Continual Learning Performance Study}\label{sec.CL.Performance}
Our goal is to decide which filters trained for intermediate task $T_t$ to prune/freeze when training the network on task $T_{t+1}$, given the sensitivity scores of layers introduced in (\ref{Def: Delta}), so that the predictive power of the network is maximally retained and not only forgetting does not degrade performance but we also gain a performance improvement. In this section, we first take an in-depth look at the layers and show the relationship between task sensitive and task useful layers. Second we provide an analysis in which we show that sensitive layers affect performance if they get frozen while training the network on the new task.
\subsection{Performance Analysis}
The motivation of our objective in this section is that the difference between the loss functions produced by the original network $F^{(L)}$ and the one produced by the frozen network $\widetilde{F}_t^{(L)}$ should be maximized with respect to sensitive and important filters. We begin this by showing that sensitive layers are useful in performance improvement of the network.
\begin{theorem}\label{thm.1}
For a given sequence of joint random variables $(\BX_t,T_t)\sim \mathcal{D}_t$ and network $F^{(L)}$, if the $l$-th layer, $f^{(l)}$ is a $t$-task-sensitive layer then it is a $t$-task-useful layer. 
\end{theorem}

\begin{theorem}\label{thm.2}
Suppose input $\mathbf{x}_t$ and label $y_t$ are samples from $(\BX_t, T_t)$ with joint distribution $\mathcal{D}_t$. For a given distribution $\mathcal{D}_t$, if the layer $l$ is a $t$-task-useful layer,
\begin{equation}\label{new}
    {\mathbb{E}}_{(\mathbf{X}_t,T_t)\sim \mathcal{D}_t}\big[T_t \cdot G_l\circ f^{(l)}(K_{l-1} \circ\mathbf{X}_t)\big]\geq \gamma_{tl},
\end{equation}
where $G_l: \mathcal{L}_l \mapsto \mathcal{T}_t$ and $K_l: \mathcal{X}_t \mapsto \mathcal{L}_l$ are map functions. Then removing layer $l$ decreases the performance i.e.
\begin{equation}\label{claim.2}
   d(F^{(L)}_t,\widetilde{F}_t^{(L)}):=  \mathbb{E}_{(\mathbf{X}_t,Y_t)\sim D_t}\left\{L_t(F^{(L)}_t(\mathbf{X}_{t}),Y_{t}) - L_t(\widetilde{F}_t^{(L)}(\mathbf{X}_{t}),Y_{t}) \right\}>K(\gamma_{tl}).
\end{equation}
Here $\widetilde{F}_t^{(L)}:=F^{(L)}_t/\widetilde{f}_{t}^{(l)}\in \mathcal{F}$ is the network with $L$ layers when layer $l$ is frozen while training on task $t$. The function $K(\gamma_{tl})$ is increasing in $\gamma_{tl}$. 
\end{theorem}
An immediate result from the combination of Theorems~\ref{thm.1} and \ref{thm.2} is stated below: 
\begin{theorem}\label{thm.3}
Suppose input $\mathbf{x}_t$ and label $y_t$ are samples from joint random variables $(\BX_t,T_t)$ with distribution $\mathcal{D}_t$. For a given distribution $\mathcal{D}_t$, if the layer $l$ is a $t$-task-sensitive layer i.e. $\Delta_{t}(f^{(l)},f^{(l+1)})\geq \eta_{tl}$, then the performance difference between $ d(F^{(L)}_t,\widetilde{F}_t^{(L)})$ is bounded as 
\begin{equation}\label{thm1:main-Eq.1}
   d(F^{(L)}_t,\widetilde{F}_t^{(L)}):=  \mathbb{E}_{(\mathbf{X}_t,Y_t)\sim D_t}\left\{L_t(F^{(L)}_t(\mathbf{X}_{t}),Y_{t}) - L_t(\widetilde{F}_t^{(L)}(\mathbf{X}_{t}),Y_{t}) \right\}\geq g(\eta_{tl}), 
\end{equation}
where $g$ is an increasing function of $\eta_{tl}$.  
Here $\widetilde{F}_t^{(L)}:=F^{(L)}_t/\widetilde{f}_{t-1}^{(l)}\in \mathcal{F}$ is the network with $L$ layers when layer $l$ is frozen while training on task $t$.
\end{theorem}
One important takeaway from this theorem is that as sensitivity between layers $\eta_{tl}$ increases the performance gap between the original and frozen network's loss functions increases. An important property of filter importance is that it is a probabilistic measure and can be computed empirically along the network. The total loss (empirical risk) on the training set for task $T_t$ is approximated by $\frac{1}{|\mathcal{T}_t|}\sum\limits_{(\bx_t,y_t)} y_t\ell_t(\omega_t;\bx_t,y_t)$, where $\ell_t$ is a differentiable loss function (\ref{loss-func}) associated with data point $(\bx_t,y_t)$ for task $T_t$ or we use cross entropy loss. 

\subsection{Forgetting Analysis}
When sequentially learning new tasks, due to restrictions on access to examples of previously seen tasks, managing the forgetting becomes a prominent challenge. 
In this section we focus on measuring the forgetting in CL with two tasks. It is potentially possible to extend these findings to more tasks.\\
Let $\omega^*_t$ and $\omega^*_{t+1}$ be the convergent parameters after training has been finished for the tasks $T_t$ and $T_{t+1}$ sequentially. Forgetting of the $t$ task is defined as 
\begin{equation}\label{definition-forgetting}
 {O}_t :=  \ell_t(\omega^*_{t+1})-\ell_t(\omega^*_{t})
\end{equation}
In this work, we propose the expected forgetting measure based on correlation between task $T_t$ and forgetting (\ref{definition-forgetting}) given distribution $\mathcal{D}_t$:
\begin{definition} 
{\rm (Expected Forgetting)} Let $\omega^*_t$ and $\omega^*_{t+1}$ be the convergent or optimum parameters after training has been finished for the $t$ and $t+1$ task sequentially. 
The expected forgetting denoted by ${EO}_t$ is defined as
\begin{equation}\label{definition-expected-forgetting}
{EO}_t :=   {\mathbb{E}}_{(\mathbf{X}_t,T_t)\sim \mathcal{D}_t}\left[T_t\cdot\big| \left(\ell_t(\omega^*_{t+1})-\ell_t(\omega^*_{t}) \right)\big|\right]. 
\end{equation}
\end{definition}
\begin{theorem}\label{thm.4}
Suppose input $\mathbf{x}_t$ and label $y_t$ are samples from joint distribution $\mathcal{D}_t$. For a given distribution $\mathcal{D}_t$, if the layer $l$ is a $t$-task-useful layer,
\begin{equation}
    {\mathbb{E}}_{(\mathbf{X}_t,Y_t)\sim \mathcal{D}_t}\big[Y_t \cdot G_l\circ f^{(l)}(K_{l-1} \circ\mathbf{X}_t)\big]\geq \gamma_{tl},
\end{equation}
then expected forgetting ${EO}_t$ defined in (\ref{definition-expected-forgetting}) is bounded by $\epsilon(\gamma_{tl})$, a decreasing function of $\gamma_{tl}$ i.e.
\begin{equation}\label{claim.2}
  {\widetilde{EO}}_t:= \mathbb{E}_{(\mathbf{X}_t,Y_t)\sim D_t}\left\{L_{t}(\widetilde{F}^{(L)}_{t+1}(\mathbf{X}_{t}),Y_{t}) - L_t({F}_t^{(L)}(\mathbf{X}_{t}),Y_{t}) \right\}< \epsilon(\gamma_{tl}),
\end{equation}
where $\widetilde{F}_{t+1}^{(L)}:=F^{(L)}_{t+1}/\widetilde{f}_{t+1}^{(l)}\in \mathcal{F}$ is the network with $L$ layers when layer $l$ is frozen while training on task $t+1$. 
\end{theorem}
A few notes on this bound: (1) based on our finding in (\ref{claim.2}), we analytically show that under the assumption that the $l$-th layer is highly  $t$-task-useful i.e. when the hyperparameter $\gamma_{tl}$ is increasing then average forgetting is decreasing if we freeze the layer $l$ during training the network on new task $T_{t+1}$. This is achieved because $\epsilon(\gamma_{tl})$ is a decreasing function with respect to $\gamma_{tl}$; (2) by a combination of Theorems~\ref{thm.1} and \ref{thm.2} we achieve an immediate result that if layer $l$ is $t$-task-sensitive then forgetting is bounded by a decreasing function of threshold $\eta_{tl}$, $\epsilon(\eta_{tl})$; (3) We prove that the amount of forgetting that a network exhibits from learning the tasks sequentially correlates with the connectivity properties of the filters in consecutive layers. In particular, the larger these connections are, the less forgetting happens. We empirically verify the relationship between forgetting and average connectivity in Section~\ref{section.experiment}.  
\subsection{A Bound on ${EO}_t$ Using Optimal Freezing Mask}
Let $\omega_t^*$ be the set of parameters when the network is trained on task $T_t$, the optimal sparsity for layer $f^{(l)}$ with optimal mask ${m^*}^{(l)}_{t+1}$ while training on task $T_{t+1}$ is achieved by 
   \begin{align}
      (\omega_{t+1}^*, {m^*}^{(l)}_{t+1}):= {argmin}_{\omega_{t+1},m} \mathbb{E}_{(\mathbf{X}_t,Y_t)\sim D_t}\left\{\big\vert Y_t\cdot\left(\ell_t({m}^{(l)}_{t+1}\odot{\omega}_{t+1})-\ell_t(\omega^*_t)\right)\big\vert\right\},
   \end{align}
   where ${m^*}^{(l)}_{t+1}$ is the binary mask matrix created after freezing filters in the $l$-th layer after training on task $T_t$ (masks are applied to the past weights) and before training on task $T_{t+1}$. Denote ${P^*}^{(l)}_m=\frac{\|{m^*}^{(l)}_{t+1}\|_0}{|{{\omega^*}^{(l)}}_{t+1}|}$ the optimal sparsity of frozen filters in layer $l$ in the original network $F^{(L)}$.
\begin{definition}
{\rm (Task-Fully-Sensitive Layer)} The $l$-th layer, $f^{(l)}$, is called a {\it $t$-task-fully-sensitive} layer if the average information flow between filters in layers $l$ and $l+1$ is maximum i.e.
\begin{equation}
  \Delta_t(f^{(l)},f^{(l+1)}):=  \frac{1}{M_{l}\; M_{l+1}}\sum\limits_{i=1}^{M_l}\sum\limits_{j=1}^{M_{l+1}} \rho\left(f^{(l)}_i,f^{(l+1)}_j|T_t\right)\rightarrow 1 \;\;\;\; (a.s.),
\end{equation}
where $\rho$ is a connectivity measure varies in $[0,1]$. 
\begin{theorem}\label{thm.5}
Suppose input $\mathbf{x}_t$ and label $y_t$ in space $\mathcal{X}_t\times \mathcal{T}_t$ are samples from random variables $(\BX_t,T_t)$ with joint distribution $\mathcal{D}_t$. For a given distribution $\mathcal{D}_t$, if layer $l$ is t-task-fully-sensitive and  ${P}^{*(l)}_m=\frac{\|{m^*}^{(l)}_{t+1}\|_0}{|{{\omega^*}^{(l)}}_{t+1}|} \rightarrow 1$ (a.s.), this means that the entire layer $l$ is frozen when training on task $T_{t+1}$. Let ${{\widetilde{\omega^*}}}^{(l)}_{t+1}$ be the optimal weight set for layer $l$, masked and trained on task $T_{t+1}$, ${{\widetilde{\omega^*}}}^{(l)}_{t+1}={m^*}^{(l)}_{t+1}\odot {\omega^*}^{(l)}_{t+1}$,
Then the expected forgetting ${\widetilde{EO}}_t$ defined in 
      \begin{align}
{\widetilde{EO}}_t= \mathbb{E}_{(\mathbf{X}_t,T_t)\sim D_t}\left\{ |T_t\cdot\left(\ell_t({\widetilde{\omega}}^*_{t+1})-\ell_t(\omega^*_t)|\right)\right\},\;\;\hbox{is bounded by}
   \end{align}
\begin{equation}\label{EO-bound}
    {\widetilde{EO}}_t \leq \frac{1}{2}\mathbb{E}_{(\mathbf{X}_t,T_t)\sim D_t}\left\{T_t\cdot {\lambda}_t^{max}\left(C+\frac{C_\epsilon}{{\lambda}_t^{max}}\right)^2\right\}, \;\;\;C,\&\;C_\epsilon\;\hbox{are constants,}
\end{equation}
and $\lambda^{max}_t$ is the maximum eigenvalue of Hessian $\nabla^2\ell_t(\omega^*_t)$. 
\end{theorem}
\end{definition}
Based on the argumentation of this section, we believe the bound found in (\ref{EO-bound}) can provide a supportive study in how freezing rate affects forgetting explicitly. In \cite{mirzadeh2020understanding}, it has been shown that lower $\lambda_t^{max}$ or equivalently wider loss function $L_t$ leads to less forgetting however, our bound in (\ref{EO-bound}) is not a monotonic function of maximum eigenvalue of Hessian. Therefore we infer that when a layer has highest connectivity, freezing the entire layer and blocking it for a specific task does not necessarily control the forgetting. 
Our inference is not only tied to the reduction of $\lambda_t^{max}$ which  describes the width of a local minima~\cite{keskar2016large}, but we also need to rely on other hidden factors that is undiscovered for us up to this time. Although we believe that to reduce forgetting, each task should push its learning towards information preservation by protecting sensitive filters and can possibly employ the same techniques used to widen the minima to improve generalization.
\subsection{Key Components to Prove Theorems }\label{Sec.key.com}
The main proofs of Theorems~\ref{thm.1}-\ref{thm.5} are provided in supplementary materials, however in this section, we describe a set of widely used key strategies and components that are used to prove findings in Section~\ref{sec.CL.Performance}.\\
\noindent{\it Theorem~\ref{thm.1}} To prove that a task-sensitive layer is a task-useful layer, we use key components:\\
{\bf (I)} Set $\overline{\sigma}_j(s)=s.\sigma_j(s)$ where $\sigma_j$ is activation function:
$$
\Delta_t(f^{(l)},f^{(l+1)})\propto\sum\limits_{i=1}^{M_l}\sum\limits_{y_t\in T_t} \pi(y_t)\mathbb{E}\left[\sum\limits_{j=1}^{M_{l+1}}\overline{\sigma}_j\left(f_i^{(l)}(\mathbf{X}_t)\right)|T_t=y_t\right].
$$
{\bf (II)} There exist a constant $C_t$ such that
\begin{align*}
C_t \sum\limits_{i=1}^{M_l}\sum\limits_{y_t\in T_t} y_t\pi(y_t){\mathbb{E}}_{\mathbf{X}_t|y_t}\left[ f^{(l)}_i(\mathbf{X}_t)|T_t=y_t\right] \nonumber\\
\geq \sum\limits_{i=1}^{M_l}\sum\limits_{y_t\in T_t} \pi(y_t)\mathbb{E}\left[\sum\limits_{j=1}^{M_{l+1}}\overline{\sigma}_j\left(f_i^{(l)}(\mathbf{X}_t)\right)|T_t=y_t\right].   
\end{align*}
\paragraph{\it Theorem~\ref{thm.2}} Let ${\omega}^*_t$ and $\widetilde{\omega}^{*}_t$ be the convergent or optimum parameters after training $F^{(L)}_t$ and $\widetilde{F}_t^{(L)}$ has been finished for task $t$, respectively. Here we establish three important components: \\
{\bf (I)} Using Taylor approximation of $\ell_t$ around $\widetilde{\omega}^*_t$:
$$
 \ell_t(\omega^*_t)-\ell_t(\widetilde{\omega}^*_t)\approx \frac{1}{2}(\omega^*_t - \widetilde{\omega}^*_t)^T \nabla^2\ell_t(\widetilde{\omega}^*_t) (\omega^*_t - \widetilde{\omega}^*_t).
$$
{\bf (II)} Let $ \widetilde{\lambda}_t^{min}$ be the minimum eigenvalue of $\nabla^2\ell_t(\widetilde{\omega}^*_t)$, we show 
\begin{align*}
 &\frac{1}{2} \mathbb{E}_{(\mathbf{X}_t,T_t)\sim D_t}\left[T_t\cdot\left( (\omega^*_t - \widetilde{\omega}^*_t)^T \nabla^2\ell_t(\widetilde{\omega}^*_t) (\omega^*_t - \widetilde{\omega}^*_t)\right)\right]\\
&\qquad\geq \frac{1}{2} \mathbb{E}_{(\mathbf{X}_t,T_t)\sim D_t}\left[T_t\cdot\left(\widetilde{\lambda}_t^{min}\| \omega^*_t - \widetilde{\omega}^*_t\|^2\right)\right].
\end{align*}
{\bf (III)} There exist a constant $C^{(l)}$ and a map function $G_l: \mathcal{L}_l \mapsto \mathcal{T}_t$ such that
\begin{align*}
   & \mathbb{E}_{(\mathbf{X}_t,T_t)\sim D_t}\left[T_t\cdot G_l \circ \sigma^{(l)}_t\big((\omega^*_t-\widetilde{\omega}^*_t)\mathbf{X}_t\big)\right]\nonumber\\
  & \quad \leq C^{(l)} \;  \mathbb{E}_{(\mathbf{X}_t,T_t)\sim D_t}\left[T_t\cdot G_l \circ \big|\sigma^{(l)}_t(\omega^*_t\mathbf{X}_t)-\sigma^{(l)}_t(\widetilde{\omega}^*_t\mathbf{X}_t) \big|\right].
\end{align*}
\paragraph{ Theorem~\ref{thm.4}} Let $\widetilde{\omega}_{t+1}^*$ be the optimal weight after training $\widetilde{F}_{t+1}^{(L)}$ on task $t+1$. Here are the key components we need to use to prove the theorem: \\
{\bf (I)} we show
$$
    \mathbb{E}_{(\mathbf{X}_t,T_t)\sim D_t}\left\{T_t\cdot\left(\ell_t(\widetilde{\omega}_{t+1}^*)-\ell_t(\widetilde{\omega}^*_t)\right)\right\}\leq \frac{1}{2}\mathbb{E}_{(\mathbf{X}_t,T_t)\sim D_t}\left\{T_t\cdot \widetilde{\lambda}_t^{max}\|\widetilde{\omega}_{t+1}^*-\widetilde{\omega}^*_t\|^2\right\},
$$
{\bf (II)} Let $\widetilde{w}'_t$ be the convergent or (near-) optimum parameters after training $\widetilde{F}^{(L)}_t$ and $\widetilde{\lambda}_t^{max}$ be the maximum eigenvalue of $\nabla^2\ell_t(\widetilde{\omega}^*_t)$:
 \begin{equation*} 
     \nabla\ell_t(\widetilde{\omega}'_t)-\nabla\ell_t(\widetilde{\omega}^*_t) \approx  \nabla^2\ell_t(\widetilde{\omega}^*_t) (\widetilde{\omega}'_t - \widetilde{\omega}^*_t) \leq \widetilde{\lambda}_t^{max}\|\widetilde{\omega}'_t-\widetilde{\omega}^*_t\|, 
    \end{equation*}
{\bf (III)} If the convergence criterion is satisfied in the $\epsilon$-neighborhood of $\widetilde{\omega}^*_t$, then 
$$
\|\widetilde{\omega}_{t+1}^*-\widetilde{\omega}^*_t\|\leq \frac{C_{\epsilon}}{\widetilde{\lambda}_t^{max}}, \;\;\;\;\; C_\epsilon=\max\{\epsilon, 2\sqrt{\epsilon}\}.
$$
\paragraph{\it Theorem~\ref{thm.5}} Denote ${{{\widetilde{\omega^*}}}^{(l)}}_{t+1}={m^*}^{(l)}_{t+1}\odot {\omega^*}^{(l)}_{t+1}$ where ${m^*}^{(l)}_{t+1}$ is the binary freezing mask for layer $l$. For the optimal weight matrix ${\widetilde{\omega}}^*_{t+1}$ with mask ${m^*}_{t+1}$, define  
$$
{\widetilde{EO}}_t= \mathbb{E}_{(\mathbf{X}_t,T_t)\sim D_t}\left\{ |T_t\cdot\left(\ell_t({\widetilde{\omega}}^*_{t+1})-\ell_t(\omega^*_t)|\right)\right\}.
$$
{\bf (I)} Once we assume that only one connection is frozen in the training process, we can use the following upper bound of the model~\cite{lee2020layer}:
\begin{align*}
    |\ell_t({\widetilde{\omega}}^*_{t+1})-\ell_t(\omega^*_{t+1})|\leq \frac{\|{\omega^*}^{(l)}_{t+1}-{{\widetilde{\omega^*}}}^{(l)}_{t+1})\|_F}{\|{\omega^*}^{(l)}_{t+1}\|_F}\prod\limits_{j=1}^L \|{\omega^*}^{(l)}_{t+1}\|_F,
\end{align*}
{\bf (II)} Under the assumption ${P^*}^{(l)}_m=\frac{\|{m^*}^{(l)}_{t+1}\|_0}{|{{\omega^*}^{(l)}}_{t+1}|} \rightarrow 1$, we show
 \begin{align*}
{\widetilde{EO}}_t&\leq \mathbb{E}_{(\mathbf{X}_t,T_t)\sim D_t}\left\{ T_t\cdot|\left(\ell_t({{\omega}}^*_{t+1})-\ell_t(\omega^*_t)|\right)\right\}.    
 \end{align*}

\section{Related Work}\label{Sec.RW}
In recent years significant interest has been given to methods for the sequential training of a single neural network on multiple tasks. One of the primary obstacles to achieving this is catastrophic forgetting, the decrease in performance observed on previously trained tasks after learning a new task. As such, overcoming catastrophic forgetting is one of the primary desiderata of CL methods. A few approaches have been taken to address this problem, including investigation of different algorithms which limit forgetting, as well as investigation into the properties of CF itself.
\paragraph{ Catastrophic Forgetting:}
The issue of catastrophic forgetting isn't new \cite{ans2000neural,mccloskey1989catastrophic}, however the popularity of deep learning methods has brought it renewed attention. Catastrophic forgetting occurs in neural networks due to the alterations of weights during the training of new tasks. This changes the network's parameters from the optimized state achieved by training on the previous task. Recent works have aimed to better understand the causes and behavior of forgetting \cite{ramasesh2020anatomy,doan2021theoretical}, as well as to learn how the specific tasks being trained influence it and to empirically study its effects \cite{goodfellow2013empirical,nguyen2019toward}. Such theoretical research into CF provides solutions to mitigate catastrophic forgetting beyond the design of the algorithm. Similarly, our investigation into the relationship between information flow and CF provides a useful tool for reducing forgetting independent of specific algorithm.
\paragraph{ Continual Learning:}
Several methods have been applied to the problem of CL. These methods generally fall into four categories: Regularization Approaches \cite{kirkpatrick2017overcoming,zenke2017continual}, Pruning-Based Approaches \cite{mallya2018packnet,wang2020learn,sokar2021spacenet}, Replay Methods \cite{shin2017continual,wu2018incremental}, and Dynamic Architectures \cite{rusu2016progressive,yoon2017lifelong}. Regularization Approaches attempt to reduce the amount of forgetting by implementing a regularization term on previously optimized weights based on their importance for performance. Replay Methods instead store or generate samples of past tasks in order to limit forgetting when training for a new task. Dynamic Architectures expand the network when new tasks are encountered in order to accommodate them. Lastly, Pruning-Based Methods aim to freeze the most important partition of weights in the network for a given task before pruning any unfrozen weights.\\

\noindent While pruning-based methods are able to remove forgetting through the use of the masking (freezing of weights), they are often implemented to make simple pruning decisions, either using fixed pruning percents for the full network or relying on magnitude-based pruning instead of approaches which utilize available structural information of the network. Other recent works have demonstrated the importance of structured pruning \cite{chen2020long,golkar2019continual}, suggesting that pruning-based CL methods would benefit from taking advantage of measures of information such as connectivity.
While these methods commonly use fixed pruning percentages across the full network, some work outside of the domain of CL which investigate different strategies for selecting layer-wise pruning percents, and together they demonstrate the importance of a less homogeneous approach to pruning \cite{lee2020layer,saha2021space}.

\section{Experimental Evidence}\label{section.experiment}
To evaluate the influence of considering knowledge of information flow when training on sequential tasks, we perform multiple experiments demonstrating improved performance when reducing pruning on the layers with high values of the measure $\Delta_t(f^{(l)},f^{(l+1)})$. The experimental results section is divided into two main parts aligning with the overall goal of analyzing downstream information across layers. The first part discusses the performance of the CL in the context of protecting highly task sensitive layers in the pruning steps of learning process by adding multiple tasks in a single neural network~\cite{mallya2018packnet}. The second part focuses on the connectivity across layers given tasks and how connectivity varies across the layers and between tasks.
\paragraph{Setting:}
We carry out training with a VGG16 model on the CIFAR10/100 dataset. We additionally perform experiments on the Permuted MNIST dataset to determine how the characteristics of information flow differ between datasets (supporting experiments on MNIST are included in the supplementary materials). Three trials were run per experiment. After training on a given task $T_t$, and prior to pruning, we calculate $\Delta_t(f^{(l)},f^{(l+1)})$ between each adjacent pair of convolutional or linear layers as in \ref{Def: Delta}. As a baseline we prune $80\%$ of the unfrozen weights in each layer (freezing the remaining $20\%$). We prune the lowest-magnitude weights.
\subsection{How Do Task Sensitive Layers Affect Performance?}
\begin{figure}[t!]
    \centering
 \begin{subfigure}[b]{0.4\textwidth}
    \includegraphics[width=\columnwidth]{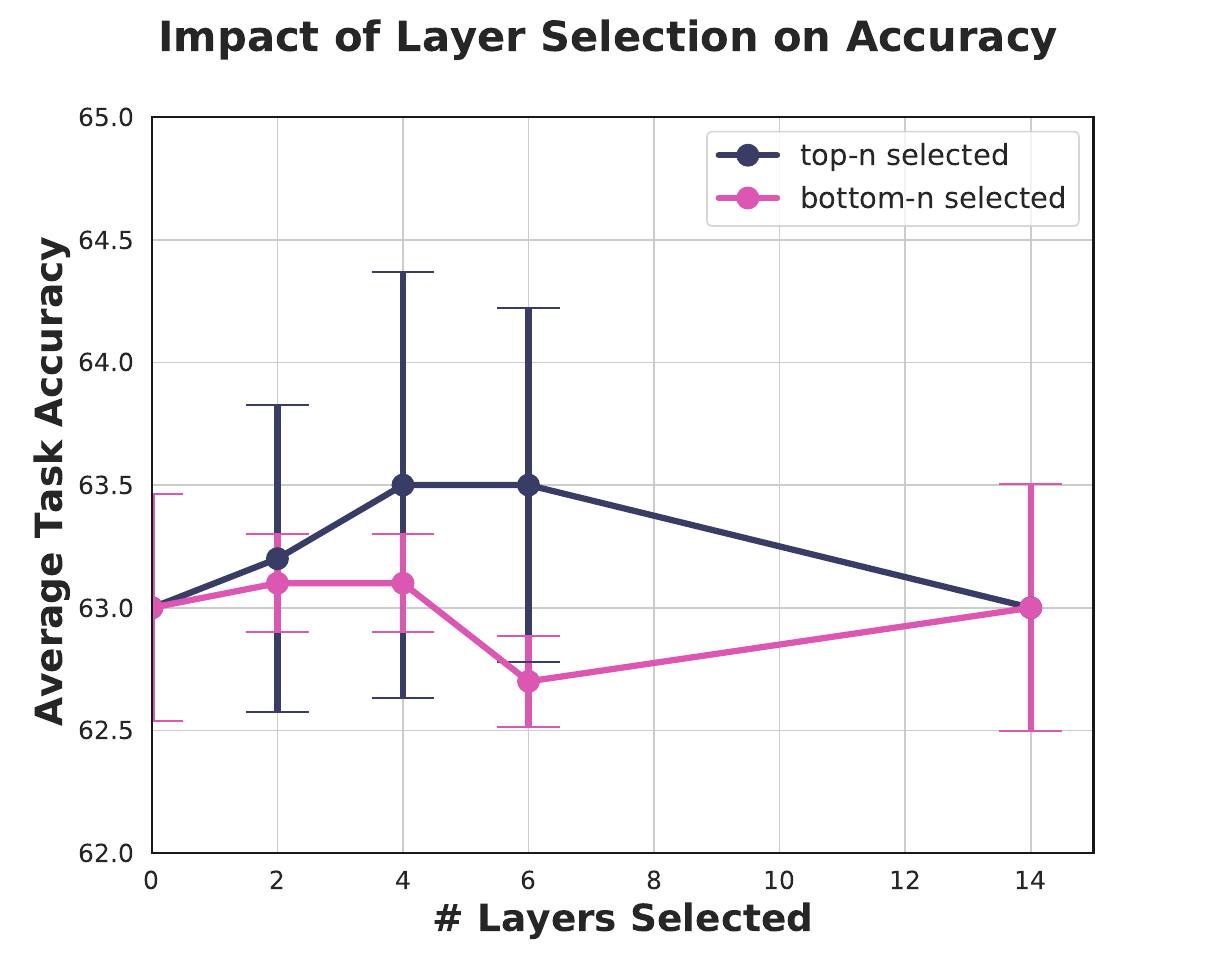}
 \end{subfigure}
 \begin{subfigure}[b]{0.4\textwidth}
    \includegraphics[width=\columnwidth]{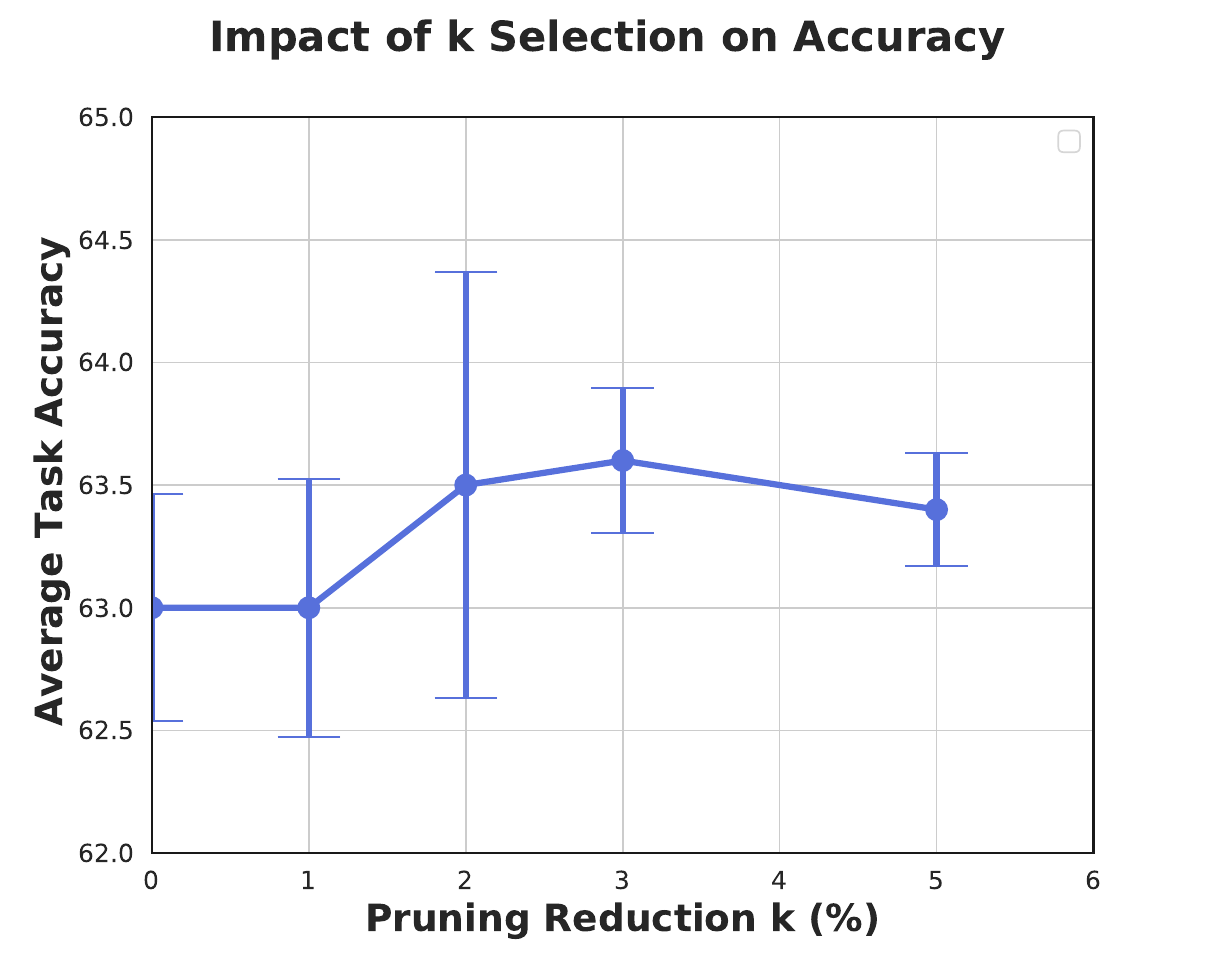}
 \end{subfigure}    
\caption{The average accuracy across tasks is reported for varying values of $n$ when $k=2\%$(left) and $k$ when $n=4$(right), where $n$ is the number of layers selected for reduced pruning and $k$ is the hyper-parameter dictating how much the pruning on selected layers is reduced by. We compare the performance when the $n$ layers are selected as the most connected layers (top-n) or least connected.}
\label{fig:performance}
\end{figure}

\paragraph{Top-Connectivity Layer Freezing:}
 For this experiment use the calculated values of $\Delta_t$. We then select the $n$ layers with the highest value of $\Delta_t$ and reduce the number of weights pruned in those layers for Task $T_t$ by $k\%$, where both $n$ and $k$ are hyper-parameters. This reduction is determined individually for each task, and only applies to the given task. By reducing pruning on the most task-sensitive layer, information flow through the network is better maintained, preserving performance on the current task. This is demonstrated in Fig. \ref{fig:performance}, in which selecting the most connected layers for reduced pruning outperforms the selection of the least connected layers. Although $n$ and $k$ have the same value in both cases, by selecting the top-n layers we better maintain the flow of information by avoiding pruning highly-connected weights. By taking values of $n>1$, we can account for cases where reducing the pruning on a single layer doesn't sufficiently maintain the flow of information above the baseline. For this reason we demonstrate the effects of changing the value of $n$. Similarly, we show the impact of varying $k$, the reduction in pruning for selected layers.
\paragraph{ Connectivity Analysis:} To better characterize our measure of information flow and determine which layers are deemed most task-sensitive, we plot the values of $\Delta_t(f^{(l)},f^{(l+1)})$ for each convolutional or linear layer, as seen in Fig. \ref{fig.freeze-connec}, Fig. \ref{fig.topn-connec}, and Fig. \ref{fig:mnistconn}. In these figures we look at how connectivity varies over several experimental setups as we change $n$ and $k$ during the selection and freezing of the top-n connected layers. We compare these trends to those seen when performing the baseline ($n,k=0$) on Permuted MNIST.


\section{Discussion}\label{Sec.Dis}
\begin{figure}[t!]
  \centering
    \includegraphics[height=0.4\columnwidth,width=0.9\columnwidth]{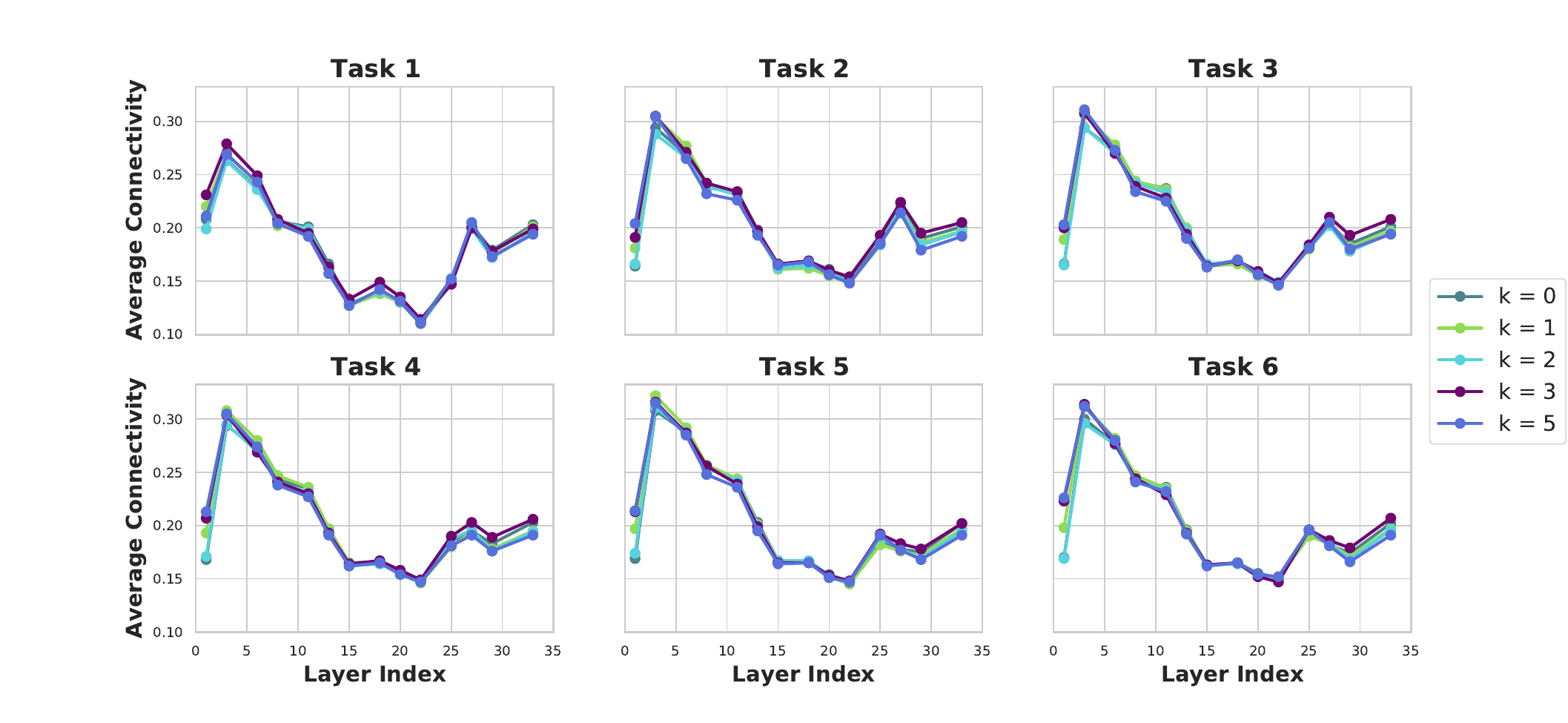}
\caption{For each layer the average connectivity value with the subsequent layer is reported. The connectivities are plotted for each task in CIFAR10/100, for various $k$ when $n=4$ most connected layers are selected for reduced pruning.}
\label{fig.freeze-connec}
\end{figure}

\begin{figure}[t!]
  \centering
    \includegraphics[height=0.42\columnwidth,width=0.9\columnwidth]{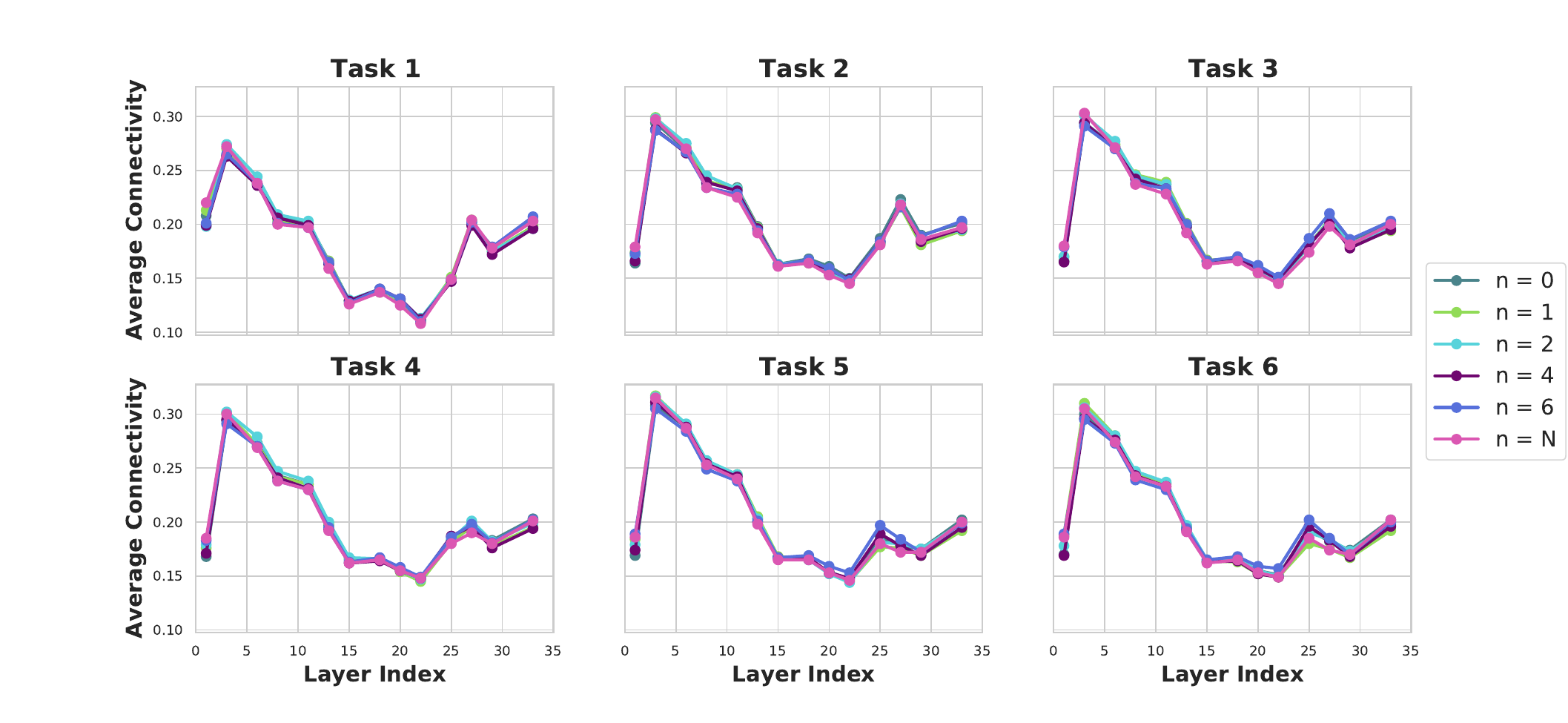}
\caption{The average connectivities  across layers with the subsequent layer is reported. The scores are plotted for each task in CIFAR10/100, when the $n$ most-connected layers are selected to have their pruning percent reduced by $k=2\%$.}
\label{fig.topn-connec}
\end{figure}

\noindent{In-depth Analysis of Bounds:} The bound established in Theorem~\ref{thm.3} shows that the performance gap between original and adapted networks with task-specific frozen layers grows as the layers contribute more in passing the information to the next layer given tasks. This gap has a direct relationship with the activations' lipschitz property and the minimum eigenvalues of Hessian at optimal weights for pruned network. From the forgetting bound in Theorem~\ref{thm.4}, we infer that as a layer is more useful for a task then freezing it reduces the forgetting more. In addition from Theorem~\ref{thm.5}, we establish that the average forgetting is a non-linear function of width of a local minima and when the entire filters of a fully sensitive layer is frozen the forgetting tends to a tighter bound. 
\begin{figure}[h!]
  \begin{center}
    \includegraphics[width=0.5\columnwidth]{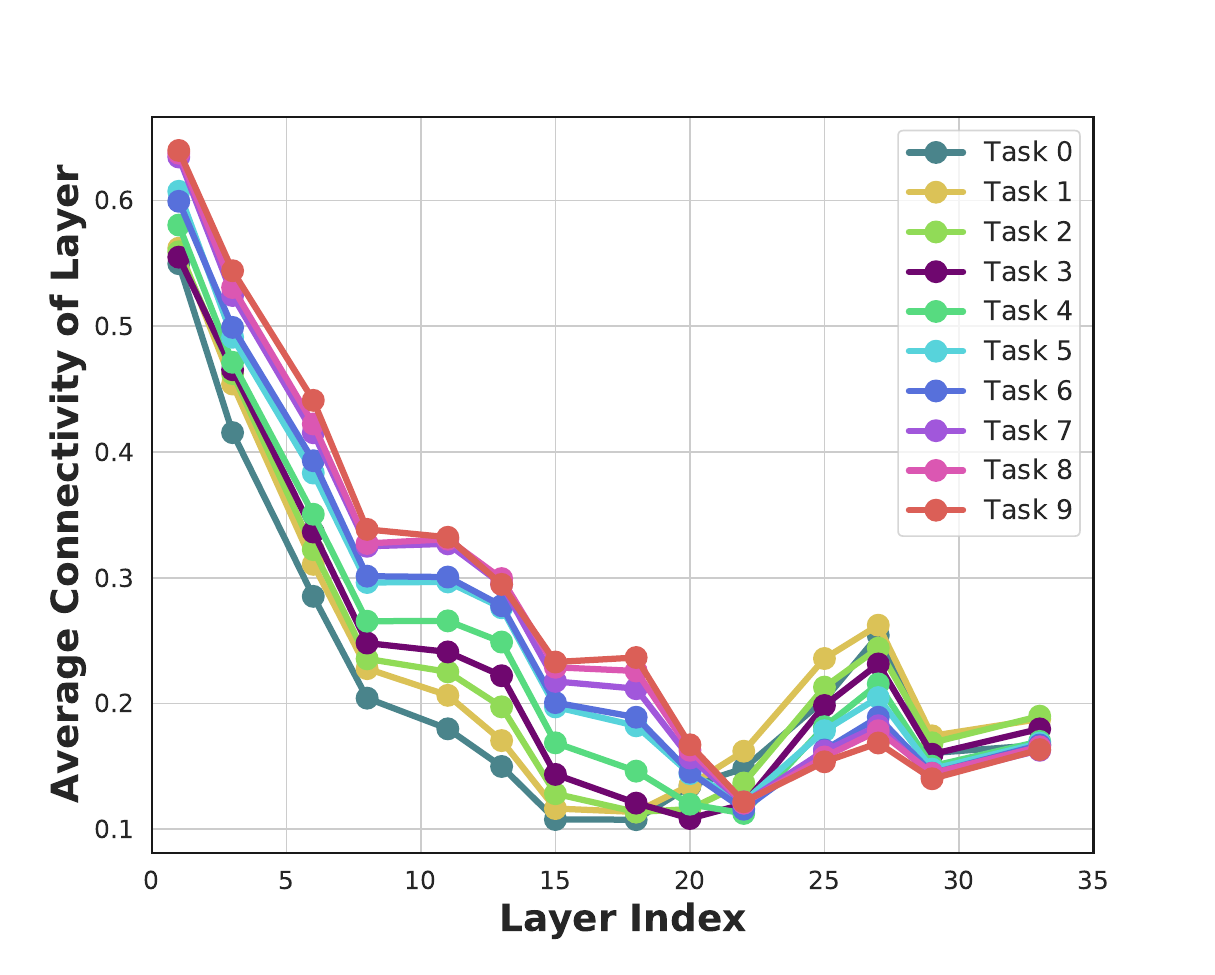}
      \end{center}
     \caption{The average connectivity for each layer is reported for training on Permuted MNIST. Training was done with the baseline setting when $n,k=0$. Each of the 10 tasks in the Permuted MNIST dataset are plotted.}
\label{fig:mnistconn}
\end{figure}
\paragraph{ Information Flow:} The connectivities plotted in Figs. \ref{fig.freeze-connec}, \ref{fig.topn-connec}, and \ref{fig:mnistconn} display patterns which remain generally consistent for a given dataset, but have noticeable differences between each dataset. For Figs. \ref{fig.topn-connec} and \ref{fig.freeze-connec} tasks 2-6, which correspond to CIFAR-100, show larger connectivities across most of the network compared to CIFAR-10, particularly in the early and middle layers. Meanwhile, for MNIST we observe connectivities which are much different from those of CIFAR-10 and CIFAR-100.
These observations suggest that when applied to different datasets, the task sensitivity of the layers in a network (VGG16 in this case) differ, indicating that the optimal pruning decision differs as well. Further, Fig. \ref{fig:mnistconn} prominently shows that as subsequent tasks are trained, the connectivity of the early layers increases while the later layers' connectivity values decrease. This can also be seen to a lesser extent in Figs. \ref{fig.freeze-connec} and \ref{fig.topn-connec}, where the peak in the last four layers decreases, while the first three layers take larger values for later tasks. This indicates that not only is the data important for determining which layers are task-sensitive, but the position of a given task in the training order is as well.\\
\noindent{Top-n Layer Freezing:} The selection of the most connected layers in Fig. \ref{fig:performance} demonstrated an improvement over the baseline, as well as over the selection of the same number of the least connected layers and the selection of all layers, showing that the improved performance isn't simply due to freezing more weights. While further work is needed to see if these results can be further improved upon, these observations lend support to the idea that making pruning decisions by utilizing knowledge of information flow in the network is an available tool to retain performance in pruning-based continual learning applications.

\section{Conclusion}\label{Sec.Con}
We've theoretically established a relationship between information flow and catastrophic forgetting and introduced new bounds on the expected forgetting. We've shown empirically how the information flow (measured by the connectivity between layers) varies between the layers of a network, as well as between tasks. Looking ahead these results highlight future possible directions of research in investigating differences in connectivity trends between various datasets, using a probabilistic connectivity measure like mutual information, and investigation on which portions of a network would be most important for passing information.

Finally, we have also empirically demonstrated that utilizing the knowledge of information flow when implementing a pruning-based CL method can improve overall performance. While these core experiments would benefit from further supporting investigations, such as the effects of different networks or tuning hyper-parameters beyond $n$ and $k$, the reported results nonetheless show promising support for the utility of information flow. Here we limited our investigation to using connectivity when determining the extent of pruning/freezing within a layer, however it would be of significant interest to see possible applications in determining which weights are pruned (as an alternative to magnitude-based pruning), or even the use of information flow in CL methods which don't utilize pruning. These are left as a very interesting future work.
%
%

\section{Acknowledgment}
This work has been partially supported by NSF 2053480; the findings are those of the authors only and do not represent any position of these funding bodies.

\renewcommand{\refname}{References}
\bibliographystyle{IEEEbib}

\bibliography{egbib}


\section{Appendices}

We first provide the full proofs of the theorems. Second we validate our analysis using the Permuted MNIST dataset. These experiments demonstrate the effects that reducing the amount pruned on highly-connected layers has on the performance of a VGG16 network and the connectivities between its layers.
\\
\\
Let us introduce the notations used in the Supplementary Material.  We are given a sequence of joint random variables $(\BX_t, T_t)$, with realization space $\CX_t\times\CT_t$ where $(\bx_t,y_t)$ is an instance of the $\CX_t\times\CT_t$ space. We assume that a given DNN has a total of $L$ layers where,
\begin{itemize}
    \item $F^{(L)}$: A function mapping the input space $\mathcal{X}$ to a set of classes $\mathcal{T}$, i.e. $F^{(L)}: \mathcal{X}\mapsto \mathcal{T}$.
    \item $f^{(l)}$: The $l$-th layer of $F^{(L)}$ with  $M_l$ as number of filters in layer $l$.
    \item $f^{(l)}_i$: $i$-th filter in layer $l$.
    \item $F^{(i,j)}:= f^{(j)}\circ\ldots \circ f^{(i)}$: A subnetwork which is a group of consecutive layers $f^{(i)},\ldots,f^{(j)}$.
    \item  $F^{(j)}:= F^{(1,j)}=f^{(j)}\circ\ldots \circ f^{(1)}$: First part of the network up to layer $j$.
    \item  $\sigma ^{(l)}$: The activation function in layer $l$.
    \item $\widetilde{f}_{t}^{(l)}$: Sensitive layer for task $t$.
    \item $\widetilde{F}_t^{(L)}:= F^{(L)}_t/\widetilde{f}_{t}^{(l)}$: The network with $L$ layers when $l$-th sensitive layer $\widetilde{f}^{(l)}$ is frozen while training on task $t$.
    \item $\pi(T_t)$: The prior probability of class label $T_t\in \mathcal{T}_t$. 
    \item $\eta_{tl}$, $\gamma_{tl}$: Thresholds for sensitivity and usefulness of $l$-th layer $f^{(l)}$ for task $t$.
    \item $\omega^{(1:i)}$: The weight matrix of subnetwork $F^{(1,i)}:= f^{(i)}\circ\ldots \circ f^{(1)}$.
\item $\omega^{(l)}$: The $l$-layer's weight matrix 
    \item $\widetilde{\omega}^{(l)}_t=m^{(l)}\odot{\omega}^{(l)}$: pruned version of the $l$-layer weight matrix.
    \item $\widetilde{w}^{(1:L)}=\left(\omega^{(1:l-1)},\widetilde{\omega}^{(l)},\omega^{(l+1,L)}\right)$: The weight matrix of network $F^{(L)}$ with pruned $l$-layer. 
    \item ${\omega}^*_t$ and $\widetilde{\omega}^{*}_t$: The convergent or optimum parameters after training $F^{(L)}_t$ and $\widetilde{F}_t^{(L)}$ has been finished for task $t$, respectively.
    \item ${{{\omega}}^*}^{(l)}_{t}$: The optimal weight set for layer $l$ and trained on task $T_{t}$.
    \item ${{\widetilde{\omega^*}}}^{(l)}_{t}$: The optimal weight set for layer $l$, masked and trained on task $T_{t}$.
    
\end{itemize}

\label{thm.sub1}
\section{Proof of Theorem~\hyperref[thm.1]{1}}
Recall Definitions~\hyperref[def.sensitivity]{1} and the Pearson correlation coefficient between the $i$-th filter in $l$-layer and $j$-th filter in $l+1$-layer defined in (\hyperref[Pearson-corr]{2}). By conditioning over task labels, the function $\rho$ becomes: 
\begin{equation}
 \rho(f_i^{(l)}, f_j^{(l+1)}|T_t)=\sum\limits_{y_t\in T_t} \pi(y_t)\mathbb{E}\left[f_i^{(l)}(\mathbf{X}_t)f_j^{(l+1)}(\BX_t)|T_t=y_t\right],
\end{equation}
where $\pi(y_t)$ is the prior probability. Thus the function $\Delta_t(f^{(l)},f^{(l+1)})$ is written as a proportion of the following term: 
\begin{align}\label{eq1-proof.1}
\Delta_t(f^{(l)},f^{(l+1)})&\propto\sum\limits_{i=1}^{M_l}\sum\limits_{j=1}^{M_{l+1}}\sum\limits_{y_t\in T_t} \pi(y_t)\mathbb{E}\left[f_i^{(l)}(\mathbf{X}_t)f_j^{(l+1)}(\mathbf{X}_t)|T_t=y_t\right]\nonumber\\
       & =\sum\limits_{i=1}^{M_l}\sum\limits_{j=1}^{M_{l+1}}\sum\limits_{y_t\in T_t} \pi(y_t)\mathbb{E}\left[f_i^{(l)}(\mathbf{X}_t)\cdot \sigma_j(f_i^{(l)}(\mathbf{X}_t))|T_t=y_t\right],
    \end{align}
where $\sigma_j$ is the activation function and $f^{(l+1)}_j(.)=\sigma_j(f^{(l)}_i(.))$. Let $\sigma_j$ be function $\overline{\sigma}_j(s)=s.\sigma_j(s)$, therefore (\ref{eq1-proof.1}) turns into:
\begin{align} \label{eq1-1-proof.1}
     &\sum\limits_{i=1}^{M_l}\sum\limits_{j=1}^{M_{l+1}}\sum\limits_{y_t\in T_t} \pi(y_t)\mathbb{E}\left[\overline{\sigma}_j\left(f_i^{(l)}(\mathbf{X}_t)\right)|T_t=y_t\right]\nonumber\\
      &=\sum\limits_{i=1}^{M_l}\sum\limits_{y_t\in T_t} \pi(y_t)\mathbb{E}\left[\sum\limits_{j=1}^{M_{l+1}}\overline{\sigma}_j\left(f_i^{(l)}(\mathbf{X}_t)\right)|T_t=y_t\right].
    \end{align}
On the other hand recall Definition~\hyperref[def-useful]{2}, the term in (\hyperref[semi-def:fromula]{3}) after conditioning on task labels with prior probabilities $\pi(y_t)$ for $y_t\in T_t$ becomes
\begin{equation}\label{eq:1-1}
    {\mathbb{E}}_{(\mathbf{X}_t,Y_t)\sim \mathcal{D}_t}\big[Y_t \cdot G_l\circ f^{(l)}(K_{l-1} \circ\mathbf{X}_t)\big]=\sum\limits_{y_t\in T_t} \pi(y_t){\mathbb{E}}_{\mathbf{X}_t|y_t}\big[y_t.G_l\circ f^{(l)}(K_{l-1} \circ\mathbf{X}_t)\big].
\end{equation}
For brevity we use $\BX_t$ for $K_{l-1} \circ\BX_t$. Let $G_l$ be a function that maps layer $f^{(l)}$ to a linear combination of filters i.e.
$$
G_l: f^{(l)}\longmapsto \sum\limits_{i=1}^{M_l} f_i^{(l)},
$$
Therefore the right hand side of (\ref{eq:1-1}) turns into 
\begin{align}
   \sum\limits_{y_t\in T_t} \pi(y_t){\mathbb{E}}_{\mathbf{X}_t|y_t}\left[\sum\limits_{i=1}^{M_l}y_t\cdot f^{(l)}_i(\mathbf{X}_t)|T_t=y_t\right]\nonumber\\
  \qquad =\sum\limits_{i=1}^{M_l}\sum\limits_{y_t\in T_t} y_t\pi(y_t){\mathbb{E}}_{\mathbf{X}_t|y_t}\left[ f^{(l)}_i(\mathbf{X}_t)|T_t=y_t\right]. 
\end{align}
Set $\beta(s):=\sum\limits_{j=1}^{M_{l+1}} \sigma_j(s)$. We know that there exists a constant $C$ such that $C\;\mathbb{E}[s]\geq \; \mathbb{E}[s.\beta(s)]$. This implies that
\begin{align}\label{eq:1-2}
C_t \sum\limits_{i=1}^{M_l}\sum\limits_{y_t\in T_t} y_t\pi(y_t){\mathbb{E}}_{\mathbf{X}_t|y_t}\left[ f^{(l)}_i(\mathbf{X}_t)|T_t=y_t\right] \nonumber\\
\geq \sum\limits_{i=1}^{M_l}\sum\limits_{y_t\in T_t} \pi(y_t)\mathbb{E}\left[\sum\limits_{j=1}^{M_{l+1}}\overline{\sigma}_j\left(f_i^{(l)}(\mathbf{X}_t)\right)|T_t=y_t\right].   
\end{align}
Note that here a possible example of $C_t=\sum\limits_{j=1}^{M_{l+1}} U_j$, where $U_j$ is an upper bound of $\sigma_j$. Notice here $y_t\geq 1$ for $y_t\in T_t$. Under the assumption that layer $f^{(l)}$ is $t$-task sensitive i.e. $\Delta_t(f^{(l)},f^{(l+1)})\geq\eta_{tl}$, the RHS of (\ref{eq:1-2}) is bounded by a proportion of $\eta_{tl}$. This combined with  (\ref{eq1-proof.1}) implies that
\begin{equation}\label{eq2-proof1}
 \sum\limits_{i=1}^{M_l}\sum\limits_{y_t\in T_t} y_t\pi(y_t){\mathbb{E}}_{\mathbf{X}_t|y_t}\left[ f^{(l)}_i(\mathbf{X}_t)|T_t=y_t\right] \geq \gamma_{tl}.  
\end{equation}
where $\gamma_{tl}\propto \eta_{tl}\big/C_t$. This concludes that layer $f^{(l)}$ is $t$-task useful. 
\label{thm.sub2}
\section{Proof of Theorem~\hyperref[thm.2]{2}}
Let ${\omega}^*_t$ and $\widetilde{\omega}^{*}_t$ be the convergent or optimum parameters after training $F^{(L)}_t$ and $\widetilde{F}_t^{(L)}$ has been finished for task $t$, respectively. In addition, training a classifier is performed by minimizing a loss function (via empirical risk minimization (ERM)) that decreases with the correlation between the weighted combination of the features and the label defined in (\hyperref[eq-loss-func2]{6}):
\begin{equation}
\mathbb{E}_{(\mathbf{X}_t,T_t)\sim D_t}\left\{L_t(F^{(L)}_t(\BX_{t}),T_{t})\right\} = \mathbb{E}_{(\mathbf{X}_t,T_t)\sim D_t}\left\{T_t\cdot \ell_t(\omega)\right\}.   
\end{equation}
Set ${\delta}_t (\omega^*_t|\widetilde{\omega}^*_t):= \ell_t(\omega^*_t)-\ell_t(\widetilde{\omega}^*_t)$. The difference between training performance of $F^{(L)}$ and  $\widetilde{F}_t^{(L)}:=F^{(L)}_t/\widetilde{f}_{t}^{(l)}\in \mathcal{F}$, the network in which layer $l$ is frozen while training on task $t$, $d(F^{(L)}_t,\widetilde{F}_t^{(L)})$, defined in (\hyperref[perform-diff]{3}) is given by
\begin{equation}\label{eq2-proof2}
 \mathbb{E}_{(\mathbf{X}_t,T_t)\sim D_t}\left[T_t\cdot{\delta}_t (\omega^*_t|\widetilde{\omega}^*_t)\right]= \mathbb{E}_{(\mathbf{X}_t,T_t)\sim D_t}\big[T_t\cdot\big(\ell_t(\omega^*_t)-\ell_t(\widetilde{\omega}^*_t)\big)\big].
\end{equation}
Using the arguments in \cite{mirzadeh2020understanding} if we write the second order Taylor approximation of $\ell_t$ around $\widetilde{\omega}^*_t$, we get
\begin{equation}
 \ell_t(\omega^*_t)\approx  \ell_t(\widetilde{\omega}^*_t) +(\omega^*_t - \widetilde{\omega}^*_t) \nabla\ell_t(\widetilde{\omega}^*_t) +\frac{1}{2}(\omega^*_t - \widetilde{\omega}^*_t)^T \nabla^2\ell_t(\widetilde{\omega}^*_t) (\omega^*_t - \widetilde{\omega}^*_t), 
\end{equation}
where $\nabla^2\ell_t(\widetilde{\omega}^*_t)$ is the Hessian for loss $\ell_t$ at $\widetilde{\omega}^*_t$. Because the model is assumed to converge to a stationary point where the gradient’s norm vanishes, $\nabla\ell_t(\widetilde{\omega}^*_t)=0$:
\begin{equation}\label{eq1-proof2}
 \ell_t(\omega^*_t)-\ell_t(\widetilde{\omega}^*_t)\approx \frac{1}{2}(\omega^*_t - \widetilde{\omega}^*_t)^T \nabla^2\ell_t(\widetilde{\omega}^*_t) (\omega^*_t - \widetilde{\omega}^*_t).
\end{equation}
We use the property that the Hessian is positive semi-definite bound (\ref{eq1-proof2}) by
\begin{equation}
 \ell_t(\omega^*_t)-\ell_t(\widetilde{\omega}^*_t)\geq \widetilde{\lambda}_t^{min}\| \omega^*_t - \widetilde{\omega}^*_t\|^2,
\end{equation}
here $ \widetilde{\lambda}_t^{min}$ is the minimum eigenvalue of $\nabla^2\ell_t(\widetilde{\omega}^*_t)$. This bounds (\ref{eq2-proof2}) by
\begin{align}
 &\frac{1}{2} \mathbb{E}_{(\mathbf{X}_t,T_t)\sim D_t}\left[T_t\cdot\left( (\omega^*_t - \widetilde{\omega}^*_t)^T \nabla^2\ell_t(\widetilde{\omega}^*_t) (\omega^*_t - \widetilde{\omega}^*_t)\right)\right]\\
&\qquad\geq \frac{1}{2} \mathbb{E}_{(\mathbf{X}_t,T_t)\sim D_t}\left[T_t\cdot\left(\widetilde{\lambda}_t^{min}\| \omega^*_t - \widetilde{\omega}^*_t\|^2\right)\right].
\end{align}
Recall that the $l$-th layer is defined as $f^{(l)}_t=\sigma^{(l)}_t\left(\omega_t \BX_t\right)$. There exists a constant $C^{(l)}$ such that 
\begin{equation}\label{eq4-proof2}
 \sigma^{(l)}_t\big((\omega^*_t-\widetilde{\omega}^*_t)\mathbf{X}_t\big)\leq C^{(l)}\big|\sigma^{(l)}_t(\omega^*_t\mathbf{X}_t)-\sigma^{(l)}_t(\widetilde{\omega}^*_t\mathbf{X}_t) \big|.
\end{equation} 
where $|.|$ is the element-wise absolute value. Next in both sides of the Ineq. (\ref{eq4-proof2})  we map $\sigma^{(l)}\in \mathcal{L}_l$ using $G_l: \mathcal{L}_l \mapsto \mathcal{T}_t$, multiple to task $T_t$ and take the expectation:
\begin{align}\label{eq6-proof2}
   & \mathbb{E}_{(\mathbf{X}_t,T_t)\sim D_t}\left[T_t\cdot G_l \circ \sigma^{(l)}_t\big((\omega^*_t-\widetilde{\omega}^*_t)\mathbf{X}_t\big)\right]\nonumber\\
  &  \leq C^{(l)} \;  \mathbb{E}_{(\mathbf{X}_t,T_t)\sim D_t}\left[T_t\cdot G_l \circ \big|\sigma^{(l)}_t(\omega^*_t\mathbf{X}_t)-\sigma^{(l)}_t(\widetilde{\omega}^*_t\mathbf{X}_t) \big|\right].
\end{align}
Given distribution $\mathcal{D}_t$, assuming that the $l$-th layer is $t$-task-useful, (\hyperref[semi-def:fromula]{3}), we have 
\begin{equation}\label{eq5-proof2}
\mathbb{E}_{(\mathbf{X}_t,T_t)\sim D_t}\left[T_t\cdot G_l \circ \sigma^{(l)}_t\big(\omega_t\mathbf{X}_t\big)\right]\geq \gamma_{tl},   
\end{equation}
Let $\omega_t=\omega^*_t-\widetilde{\omega}^*_t$ in (\ref{eq5-proof2}) and combined with (\ref{eq6-proof2}), we get
\begin{gather}\label{eq0-lemma}
 \mathbb{E}_{(\mathbf{X}_t,T_t)\sim D_t}\left[T_t\cdot G_l \circ \big|\sigma^{(l)}_t(\omega^*_t\mathbf{X}_t)-\sigma^{(l)}_t(\widetilde{\omega}^*_t\mathbf{X}_t) \big|\right]\geq \widetilde{\gamma}_{tl},
\end{gather}
where $\widetilde{\gamma}_{tl}={\gamma}_{tl}\big/C^{(l)}$. We assume that the activation $\sigma^{(l)}$ is Lipschitz continuous since it is generally true for most of the commonly used activations in neural networks such as Identity, ReLU, sigmoid, tanh, PReLU, etc. Then we know for for any $\mathbf{z},\mathbf{s}$, there exist a constant $C^{(l)}_\sigma$ such that
\begin{equation*}
    |\sigma^{(l)}(\mathbf{z})-\sigma^{(l)}(\mathbf{s})|\leq C^{(l)}_\sigma |\mathbf{z}-\mathbf{s}|.
\end{equation*}
Then it is easy to see that
\begin{equation}\label{eq-Lipschitz}
    \big|\sigma^{(l)}_t(\omega^*_t\mathbf{X}_t)-\sigma^{(l)}_t(\widetilde{\omega}^*_t\mathbf{X}_t) \big| \leq C^{(l)}_\sigma |\omega^*_t- \widetilde{\omega}^*_t||\mathbf{X}_t|,
\end{equation}
We first apply two map functions $G_l$ and $\overline{G}_l$ on left and right sides of the above inequality to map them to the space $\mathcal{T}_t$, second we multiply $T_t$, and take expectation with respect to distribution $\mathcal{D}_t$:
\begin{gather}\label{eq-Lipschitz-expectation}
\mathbb{E}_{(\mathbf{X}_t,T_t)\sim D_t}\left[T_t\cdot G_l \circ \big|\sigma^{(l)}_t(\omega^*_t\mathbf{X}_t)-\sigma^{(l)}_t(\widetilde{\omega}^*_t\mathbf{X}_t) \big|\right]\\
\leq C^{(l)}_\sigma \; \mathbb{E}_{(\mathbf{X}_t,T_t)\sim D_t}\left[T_t\cdot \overline{G}_l \circ |\omega^*_t- \widetilde{\omega}^*_t||\mathbf{X}_t|\right].
\end{gather}
Note that $\omega^*_t$ and ${\widetilde{\omega}}^*_t$ are trained weight matrix from layers 1 to $l$ with layer $l$ included and excluded in training respectively. Combining (\ref{eq0-lemma}), (\ref{eq-Lipschitz}), and (\ref{eq-Lipschitz-expectation}), we have 
\begin{equation}
    \widetilde{\gamma}_{tl} \leq C^{(l)}_\sigma \; \mathbb{E}_{(\mathbf{X}_t,T_t)\sim D_t}\left[T_t\cdot \overline{G}_l \circ|\omega^*_t- \widetilde{\omega}^*_t||\mathbf{X}_t|\right].
\end{equation}
Since $|\mathbf{X}_t|$ is bounded, there exist a constant $C_x$ such that $|\mathbf{X}_t|\leq C_x$. Thus, we have
\begin{equation}
   C_\gamma \leq \mathbb{E}_{(\mathbf{X}_t,T_t)\sim D_t}\left[T_t\cdot \overline{G}_l \circ|\omega^*_t- \widetilde{\omega}^*_t|\right]. 
\end{equation}
where $C_\gamma=\widetilde{\gamma}_{tl}\big/C_x C^{(l)}_\sigma$. Let $\overline{G}_l: |\omega^*_t- \widetilde{\omega}^*_t| \mapsto \widetilde{\lambda}_t^{min}\|\omega^*_t- \widetilde{\omega}^*_t\|^2$. 
This implies 
\begin{align}\label{eq3-lemma}
K(\gamma_{tl})
\leq \mathbb{E}_{(\mathbf{X}_t,T_t)\sim D_t}\left[T_t\cdot\left(\widetilde{\lambda}_t^{min}\| \omega^*_t - \widetilde{\omega}^*_t\|^2\right)\right],
\end{align}
is lower bounded by $K(\gamma_{tl})\propto \gamma_{tl}\big/(C^{(l)}C_xC^{(l)}_\sigma$ which is an increasing function of $\gamma_{tl}$. This concludes the proof and shows that in (\ref{eq2-proof2}), 
\begin{equation*}
   d(F^{(L)}_t,\widetilde{F}_t^{(L)})= \mathbb{E}_{(\mathbf{X}_t,T_t)\sim D_t}\left[T_t\cdot{\delta}_t (\omega^*_t|\widetilde{\omega}^*_t)\right] \geq K(\gamma_{tl}).
\end{equation*}
\label{thm.sub4}
\section{Proof of Theorem~\hyperref[thm.4]{4}}
Let $\widetilde{\omega}_{t+1}^*$ be the optimal weight after training $\widetilde{F}_{t+1}^{(L)}$ on task $t+1$. Following the arguments and notations in the proof of Theorem~\hyperref[thm.2]{2}:
\begin{equation}\label{eq-thm2-1-1}
   \mathbb{E}_{(\mathbf{X}_t,T_t)\sim D_t}\left\{L_{t}(\widetilde{F}^{(L)}_{t+1}(\mathbf{X}_{t}),T_{t}) - L_t({F}_t^{(L)}(\mathbf{X}_{t}),T_{t}) \right\}
   =\mathbb{E}_{(\mathbf{X}_t,T_t)\sim D_t}\left\{T_t\cdot\left(\ell_t(\widetilde{\omega}_{t+1}^*)-\ell_t(\omega^*_t)\right)\right\}.
\end{equation}
Subtract and add the term $\ell_t(\widetilde{\omega}^*_t)$ in (\ref{eq-thm2-1-1}): 
\begin{equation}\begin{array}{cl}\label{eq-thm2-2}
\mathbb{E}_{(\mathbf{X}_t,T_t)\sim D_t}\left\{T_t\cdot\left(\ell_t(\widetilde{\omega}_{t+1}^*)-\ell_t(\widetilde{\omega}^*_t)\right)+\left(\ell_t(\widetilde{\omega}^*_t)-\ell_t(\omega^*_t)\right)\right\} \\[10pt]
=\mathbb{E}_{(\mathbf{X}_t,T_t)\sim D_t}\left\{T_t\cdot\left(\ell_t(\widetilde{\omega}_{t+1}^*)-\ell_t(\widetilde{\omega}^*_t)\right)\right\}+\mathbb{E}_{(\mathbf{X}_t,T_t)\sim T_t}\left\{\left(\ell_t(\widetilde{\omega}^*_t)-\ell_t(\omega^*_t)\right)\right\}.
\end{array}\end{equation}
Using the arguments in \cite{mirzadeh2020understanding} we know that
\begin{equation}\label{ineq:0-1}
    \mathbb{E}_{(\mathbf{X}_t,T_t)\sim D_t}\left\{T_t\cdot\left(\ell_t(\widetilde{\omega}_{t+1}^*)-\ell_t(\widetilde{\omega}^*_t)\right)\right\}\leq \frac{1}{2}\mathbb{E}_{(\mathbf{X}_t,T_t)\sim D_t}\left\{T_t\cdot \widetilde{\lambda}_t^{max}\|\widetilde{\omega}_{t+1}^*-\widetilde{\omega}^*_t\|^2\right\},
\end{equation}
where $\widetilde{\lambda}_t^{max}$ is maximum eigenvalue of $\nabla^2\ell_t(\widetilde{\omega}^*_t)$. Let $\widetilde{w}'_t$ be the convergent or (near-) optimum parameters after training $\widetilde{F}^{(L)}_t$ has been finished for the first task. Then
\begin{equation}\label{eq:0-2-proof4}
\|\widetilde{\omega}_{t+1}^*-\widetilde{\omega}^*_t\|\leq \|\widetilde{\omega}_{t+1}^*-\widetilde{\omega}'_t\|+\|\widetilde{\omega}'_t-\widetilde{\omega}^*_t\|.     
\end{equation}
Since $\|\widetilde{\omega}_{t+1}^*-\widetilde{\omega}'_t\|$ is a constant, say $C$, we only need to bound $\|\widetilde{\omega}'_t-\widetilde{\omega}^*_t\|$. Consider two different convergence criterion: 
\begin{itemize}
    \item $\ell_t(\widetilde{\omega}'_t)-\ell_t(\widetilde{\omega}^*_t)\leq \epsilon$: We write the second order Taylor approximation of $\ell_t$ around $\widetilde{\omega}^*_t$: 
    \begin{equation}
     \ell_t(\widetilde{\omega}'_t)-\ell_t(\widetilde{\omega}^*_t) \approx \frac{1}{2}(\widetilde{\omega}'_t - \widetilde{\omega}^*_t)^T \nabla^2\ell_t(\widetilde{\omega}^*_t) (\widetilde{\omega}'_t - \widetilde{\omega}^*_t)\leq \frac{1}{2} \widetilde{\lambda}_t^{max}\|\widetilde{\omega}'_t-\widetilde{\omega}^*_t\|^2,  
    \end{equation}
    where $\widetilde{\lambda}_t^{max}$ is the maximum eigenvalue of $\nabla^2\ell_t(\widetilde{\omega}^*_t)$. Hence the convergence criterion can be written as $\frac{1}{2} \widetilde{\lambda}_t^{max}\|\widetilde{\omega}'_t-\widetilde{\omega}^*_t\|^2\leq \epsilon$, equivalently 
    \begin{equation}\label{eq1-proof4}
    \|\widetilde{\omega}'_t-\widetilde{\omega}^*_t\|^2\leq \frac{2\sqrt{\epsilon}}{\widetilde{\lambda}_t^{max}}
    \end{equation}
    \item $\nabla^2\ell_t(\widetilde{w}'_t)\leq \epsilon$: Write the first order Taylor approximation of $\nabla \ell_t$ around $\widetilde{\omega}^*_t$: 
    \begin{equation}
     \nabla\ell_t(\widetilde{\omega}'_t)-\nabla\ell_t(\widetilde{\omega}^*_t) \approx  \nabla^2\ell_t(\widetilde{\omega}^*_t) (\widetilde{\omega}'_t - \widetilde{\omega}^*_t) \leq \widetilde{\lambda}_t^{max}\|\widetilde{\omega}'_t-\widetilde{\omega}^*_t\|. 
    \end{equation}
    Hence the convergence criterion can be written as $\widetilde{\lambda}_t^{max}\|\widetilde{\omega}'_t-\widetilde{\omega}^*_t\|\leq \epsilon$, equivalently 
    \begin{equation}\label{eq2-proof4}
    \|\widetilde{\omega}'_t-\widetilde{\omega}^*_t\|\leq \frac{{\epsilon}}{\widetilde{\lambda}_t^{max}}.
    \end{equation}
\end{itemize}
Denote $C_\epsilon=\max\{\epsilon, 2\sqrt{\epsilon}\}$. Combining   (\ref{eq1-proof4}) and (\ref{eq2-proof4}) we get the LHS of (\ref{ineq:0-1}) bounded by
\begin{equation}\label{upper-bound-thm-1}
 \frac{1}{2}\mathbb{E}_{(\mathbf{X}_t,T_t)\sim D_t}\left\{T_t\cdot \widetilde{\lambda}_t^{max}\left(C+\frac{C_\epsilon}{\widetilde{\lambda}_t^{max}}\right)^2\right\}.
\end{equation}
Recalling Theorem~\ref{thm.2}, we obtain
\begin{equation}\label{upper-bound-thm-2}
  \mathbb{E}_{(\mathbf{X}_t,T_t)\sim D_t}\left\{T_t\cdot\left(\ell_t(\widetilde{\omega}^*_t)-\ell_t(\omega^*_t)\right)\right\}\leq -K(\gamma_{tl}),
\end{equation}
where $K(\gamma_{tl})$ is an increasing function of $\gamma_{tl}$. 
Combining (\ref{upper-bound-thm-1}) and (\ref{upper-bound-thm-2}) in (\ref{eq-thm2-2}), we have 
\begin{equation}\label{eq-theorem22-1}
   \mathbb{E}_{(\mathbf{X}_t,T_t)\sim D_t}\left\{T_t\cdot\left(\ell_t(\widetilde{\omega}_{t+1}^*)-\ell_t(\omega^*_t)\right)\right\} \leq \frac{1}{2}\mathbb{E}_{(\mathbf{X}_t,T_t)\sim D_t}\left\{T_t\cdot \widetilde{\lambda}_t^{max}\left(C+\frac{C_\epsilon}{\widetilde{\lambda}_t^{max}}\right)^2\right\} - K(\gamma_{tl}) , 
\end{equation}
By setting $\epsilon(\widetilde{\lambda}_t^{max},\gamma_{tl})$ the RHS of (\ref{eq-theorem22-1}), the function $\epsilon$ is a decreasing function of $\gamma_{tl}$ for given $\widetilde{\lambda}_t^{max}$ and the theorem is proved. 

\label{thm.sub5}
\section{Proof of Theorem~\hyperref[thm.5]{5}}
Recall
\begin{itemize}
\item $\omega^{(1:i)}$: The weight matrix of subnetwork $F^{(1,i)}:= f^{(i)}\circ\ldots \circ f^{(1)}$.
\item $\omega^{(l)}$: The $l$-layer's weight matrix 
    \item $\widetilde{\omega}^{(l)}_t=m^{(l)}\odot{\omega}^{(l)}$: pruned version of the $l$-layer weight matrix.
    \item $\widetilde{w}^{(1:L)}=\left(\omega^{(1:l-1)},\widetilde{\omega}^{(l)},\omega^{(l+1,L)}\right)$: The weight matrix of network $F^{(L)}$ with pruned $l$-layer. 
\end{itemize}
The optimal weight matrix ${\widetilde{\omega}}^*_{t+1}$ with mask ${m^*}_{t+1}$: 
      \begin{align}\label{eq1-proof5}
{\widetilde{EO}}_t= \mathbb{E}_{(\mathbf{X}_t,T_t)\sim D_t}\left\{ |T_t\cdot\left(\ell_t({\widetilde{\omega}}^*_{t+1})-\ell_t(\omega^*_t)|\right)\right\}.
   \end{align}
 By adding and subtracting term $\ell_t(\omega^*_{t+1})$ in (\ref{eq1-proof5}), we bound ${EO}_t$ by
 \begin{align}\label{eq4-proof5}
{\widetilde{EO}}_t&\leq \mathbb{E}_{(\mathbf{X}_t,T_t)\sim D_t}\left\{ T_t\cdot|\left(\ell_t({\widetilde{\omega}}^*_{t+1})-\ell_t(\omega^*_{t+1})|\right)\right\} \\
&\qquad+ \mathbb{E}_{(\mathbf{X}_t,T_t)\sim D_t}\left\{ T_t\cdot|\left(\ell_t({{\omega}}^*_{t+1})-\ell_t(\omega^*_t)|\right)\right\}.    
 \end{align}
 Once we assume that only one connection is frozen in the training process, we can use the following upper bound of the model~\cite{lee2020layer}:
\begin{align}
    |\ell_t({\widetilde{\omega}}^*_{t+1})-\ell_t(\omega^*_{t+1}|\leq \frac{\|{\omega^*}^{(l)}_{t+1}-{{\widetilde{\omega^*}}}^{(l)}_{t+1})\|_F}{\|{\omega^*}^{(l)}_{t+1}\|_F}\prod\limits_{j=1}^L \|{\omega^*}^{(l)}_{t+1}\|_F,
\end{align}
where $\|.\|_F$ is Frobenius norm. Here ${{\widetilde{\omega^*}}}^{(l)}_{t+1}$ is the optimal weight set for layer $l$, masked and trained on task $T_{t+1}$, ${{\widetilde{\omega^*}}}^{(l)}_{t+1}={m^*}^{(l)}_{t+1}\odot {\omega^*}^{(l)}_{t+1}$, therefore
$$
\|{\omega^*}^{(l)}_{t+1}-{{\widetilde{\omega^*}}}^{(l)}_{t+1}\|_F = \|{\omega^*}^{(l)}_{t+1}-{m^*}^{(l)}_{t+1}\odot {\omega^*}^{(l)}_{t+1}\|_F= \|{\omega^*}^{(l)}_{t+1}\left(\mathbbm{1}-{m^*}^{(l)}_{t+1}\right)\|_F.
$$
The assumption ${P^*}^{(l)}_m=\frac{\|{m^*}^{(l)}_{t+1}\|_0}{|{{\omega^*}^{(l)}}_{t+1}|} \rightarrow 1$ (a.s.) is equivalent to $\left(\mathbbm{1}-{m^*}^{(l)}_{t+1}\right)\rightarrow 0$ (a.s.). Therefore,
\begin{align}
     |\ell_t({\widetilde{\omega}}^*_{t+1})-\ell_t(\omega^*_{t+1}) | \longrightarrow 0\;\;\; (a.s.)
\end{align}
This implies that the term in the RHS of (\ref{eq4-proof5}) convergences to zero. 
Now from (\ref{eq2-proof1}) in the proof of Theorem~\hyperref[thm.1]{1}, if the $l$-th layer is fully sensitive i.e. $\eta_{tl}\rightarrow 1$ then $\gamma_{tl}\propto 1/C_t$, where $C_t$ is constant. Next using analogous arguments in (\ref{ineq:0-1})-(\ref{upper-bound-thm-1}) in the proof of Theorem~\ref{thm.4}, we have  
\begin{align}
     &\mathbb{E}_{(\mathbf{X}_t,T_t)\sim D_t}\left\{ T_t\cdot|\left(\ell_t({{\omega}}^*_{t+1})-\ell_t(\omega^*_t)|\right)\right\} \\
    & \quad\leq \frac{1}{2}\mathbb{E}_{(\mathbf{X}_t,T_t)\sim D_t}\left\{T_t\cdot {\lambda}_t^{max}\left(C+\frac{C_\epsilon}{{\lambda}_t^{max}}\right)^2\right\},
\end{align}
where $\lambda^{max}_t$ is the maximum eigenvalue of $\nabla^2\ell_t(\omega^*_t)$. Here $C$ and $C_\epsilon$ are constants.

\label{Def: Delta-equ}
\section{Further Experiments}
\subsection{Experimental Setup}
For the experiments outlined here we use a VGG16 model on the Permuted MNIST dataset to determine how the characteristics of information flow differ from what we observed on CIFAR-10/100. After training on a given task $T_t$, and prior to pruning, we calculate $\Delta_t(f^{(l)},f^{(l+1)})$ between each adjacent pair of convolutional or linear layers as in Definition~\hyperref[Def: Delta]{1}, (\hyperref[Def: Delta-equ]{1}). As a baseline we prune $80\%$ of the unfrozen weights in each layer (freezing the remaining $20\%$), pruning the lowest-magnitude weights. We run a single trial for each experiment.

\begin{figure}[t!]
    \centering
 \begin{subfigure}[b]{0.4\textwidth}
    \includegraphics[width=\columnwidth]{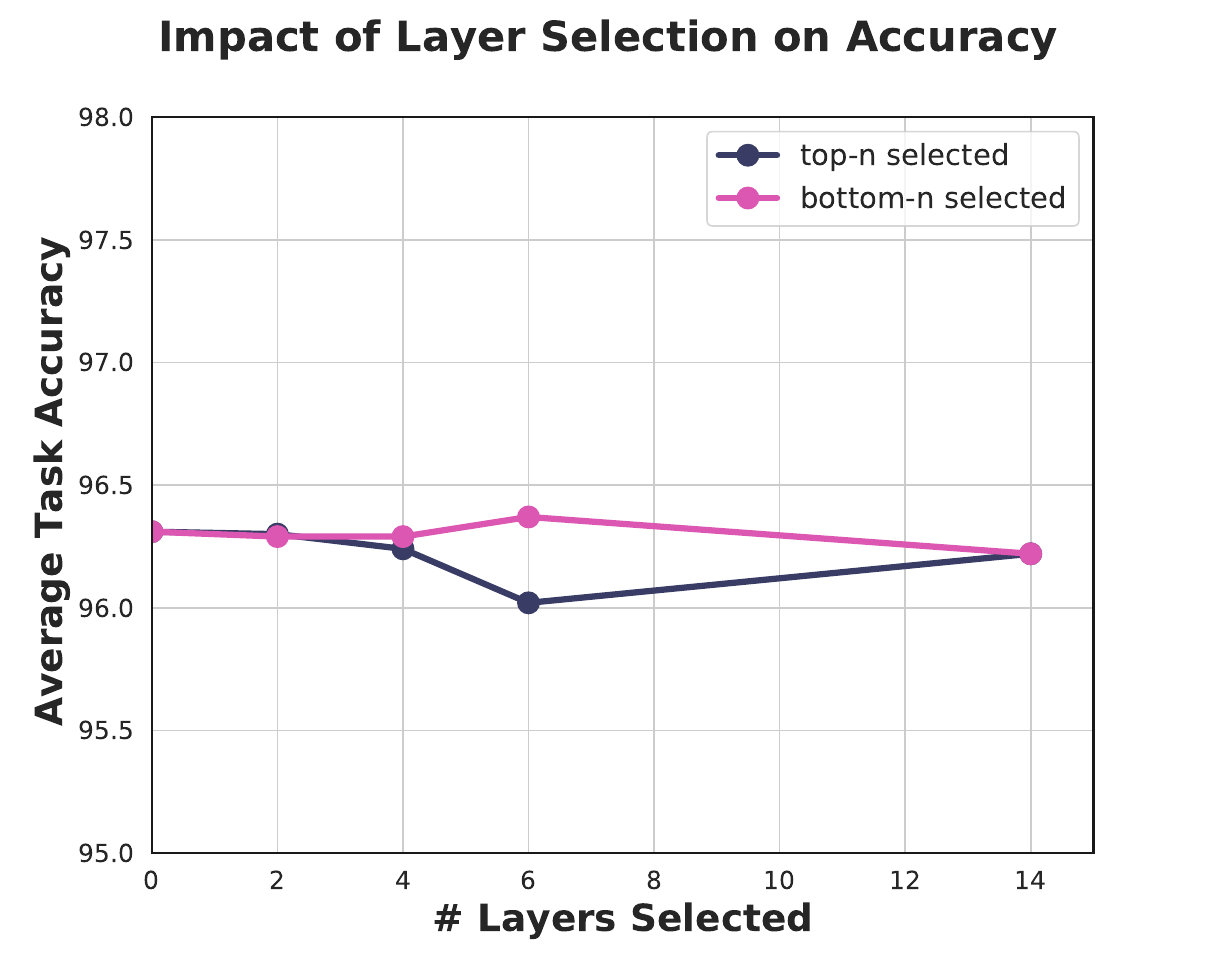}
 \end{subfigure}
 \begin{subfigure}[b]{0.4\textwidth}
    \includegraphics[width=\columnwidth]{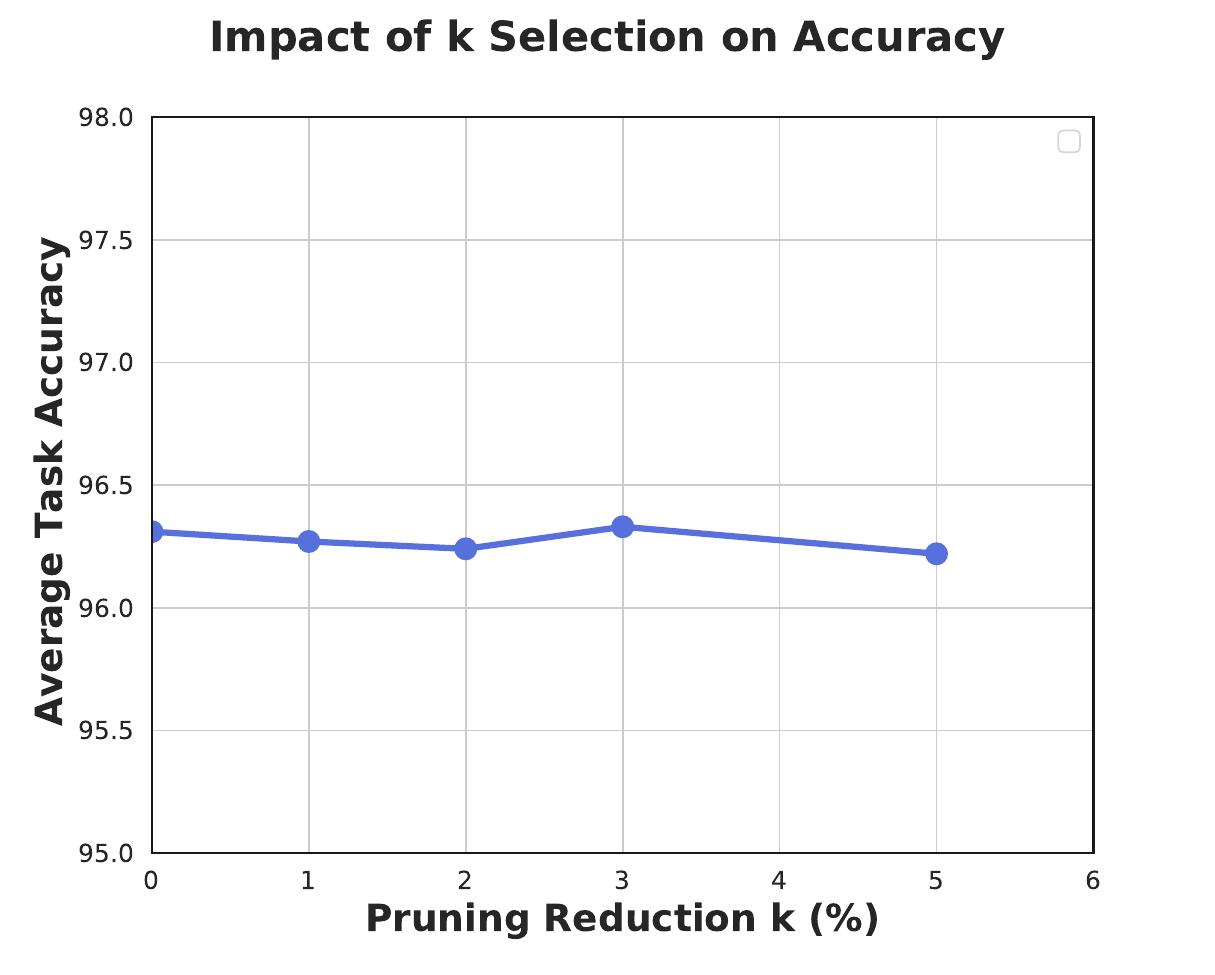}
 \end{subfigure}    
\caption{The average accuracy across tasks of Permuted MNIST is reported for varying values of $n$ when $k=2\%$ (left) and $k$ when $n=4$ (right), where $n$ is the number of layers selected for reduced pruning and $k$ is the hyper-parameter dictating how much the pruning on selected layers is reduced by. We compare the performance when the $n$ layers are selected as the most connected layers (top-n) or least connected.}
\label{figsm:performance}
\end{figure}

\begin{figure}[t!]
  \centering
    \includegraphics[height=0.42\columnwidth,width=1\columnwidth]{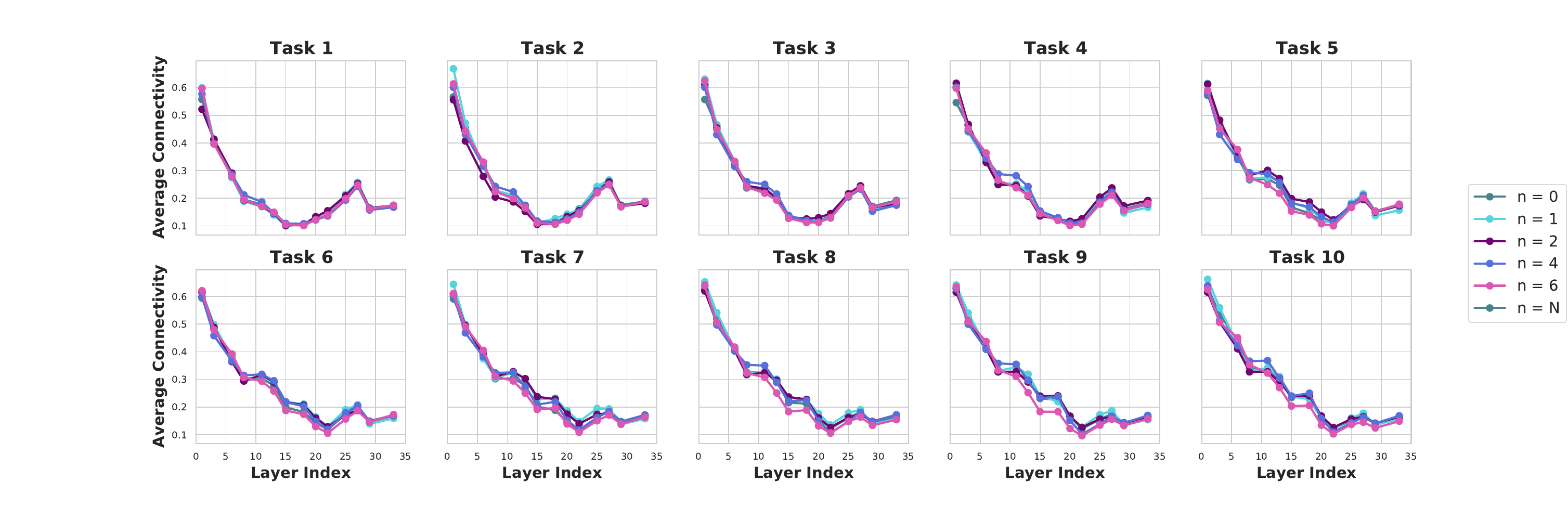}
\caption{The average connectivities  across layers with the subsequent layer is reported. The scores are plotted for each task in Permuted MNIST, when the $n$ most-connected layers are selected to have their pruning percent reduced by $k=2\%$.}
\label{figsm.topn-connec}
\end{figure}

\begin{figure}[h!]
  \centering
    \includegraphics[height=0.4\columnwidth,width=1\columnwidth]{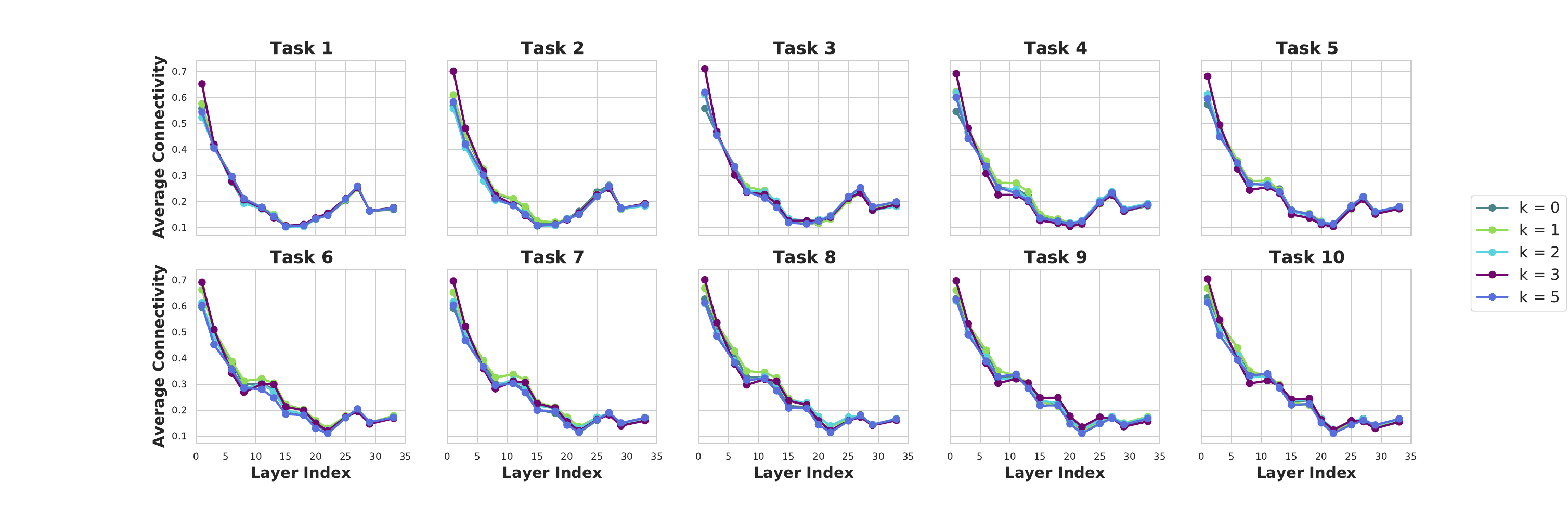}
\caption{For each layer the average connectivity value with the subsequent layer is reported. The connectivities are plotted for each task in Permuted MNIST, for various $k$ when $n=4$ most connected layers are selected for reduced pruning.}
\label{figsm.freeze-connec}
\end{figure}

\subsection{Permuted MNIST Experiment}
In this section we provide the results of implementing our experiments on the Permuted MNIST dataset. For these experiments we vary the hyper-parameters $n$ and $k$, where $n$ is the number of layers selected for reduced pruning and $k$ is the percent of pruning reduction provided to the selected layers. By varying $n$ we can see that unlike our observations with CIFAR-10/100, the overall performance remains roughly the same as the baseline (Fig. \ref{figsm:performance}) and selecting the bottom-n layers performed as well as or better than selecting the top-n. To provide an empirical observation of information downstream in the network's layers, we report the differences in connectivities for these experiments as well in Fig. \ref{figsm.topn-connec}. Similarly to our findings with CIFAR-10/100, the changes to $n$ don't appear to substantially alter the patterns in connectivity between layers that we observe throughout the Permuted MNIST. As with CIFAR-10/100, we can see the trends in subsequent tasks where the early layers' connectivities increase throughout the tasks and the later layers decrease. 

We demonstrate the effects of altering $k$ on performance with Permuted MNIST in \ref{figsm:performance}. Once again, unlike with CIFAR-10/100 we observe similar or slightly lower performance compared to the baseline which decreases as the percent continues to increase. The effects of varying $k$ on connectivity can be seen in Fig. \ref{figsm.freeze-connec}.
The results we observe with Permuted MNIST closely match those seen from CIFAR-10/100 for connectivities but not performances, which indicates that although we observe similar patterns across experiments and tasks for the two datasets, additional steps or a more systematic freezing approach may need to be established to optimally apply our knowledge of information flow to different models. In addition, this observation provides an experimental evidence that hyperparameters $\gamma_{tl}$ and $\eta_{tl}$ are data dependent.



\end{document}